\def\paperlanguage{} 
\pdfoutput=1 

\documentclass[letterpaper, 10 pt, conference]{ieeeconf}  

\usepackage{bm}
\usepackage{cite}
\usepackage{siunitx}
\usepackage{multirow}
\include{preamble}
\newcommand{\ctext}[1]{\raise0.2ex\hbox{\textcircled{\scriptsize{#1}}}}

\IEEEoverridecommandlockouts                              

\title{\LARGE \textbf
  {
    \switchlanguage%
    {%
      Body Design and Gait Generation of \\Chair-Type Asymmetrical Tripedal Low-rigidity Robot
    }%
    {%
      椅子型非対称三脚移動ロボットの身体設計と歩容生成
    }%
  }
}

\author{Shintaro Inoue$^{1}$, Kento Kawaharazuka$^{1}$, Kei Okada$^{1}$, and Masayuki Inaba$^{1}$
  \thanks{$^{1}$ The authors are with the Department of Mechano-Informatics, Graduate School of Information Science and Technology, The University of Tokyo, 7-3-1 Hongo, Bunkyo-ku, Tokyo, 113-8656, Japan.
    {\texttt\small [s-inoue, kawaharazuka, okada, inaba]@jsk.t.u-tokyo.ac.jp}
  }
}
\begin{document}

\maketitle
\thispagestyle{empty}
\pagestyle{empty}

\begin{abstract}
\switchlanguage%
{%
  In this study, a chair-type asymmetric tripedal low-rigidity robot was designed based on the three-legged chair character in the movie ``Suzume'' and its gait was generated.
  Its body structure consists of three legs that are asymmetric to the body, so it cannot be easily balanced.
  In addition, the actuator is a servo motor that can only feed-forward rotational angle commands and the sensor can only sense the robot's posture quaternion.
  In such an asymmetric and imperfect body structure, 
  we analyzed how gait is generated in walking and stand-up motions by generating gaits with two different methods:
  a method using linear completion to connect the postures necessary for the gait discovered through trial and error using the actual robot,
  and a method using the gait generated by reinforcement learning in the simulator and reflecting it to the actual robot.
  Both methods were able to generate gait that realized walking and stand-up motions, 
  and interesting gait patterns were observed, which differed depending on the method, and were confirmed on the actual robot.
  Our code and demonstration videos are available here: https://github.com/shin0805/Chair-TypeAsymmetricalTripedalRobot.git
}%
{%
  本研究では，2022年に公開された映画「すずめの戸締まり」に登場する三本脚で動く椅子のキャラクターをモチーフに，
  椅子型非対称三脚ロボットを設計し，歩容を生成した．
  その身体構造は，体に対して非対称な3本脚で構成されており，一つの脚は2自由度をもち，前後左右に脚を動かすことができる．
  一方で，アクチュエータは回転角度指令しかできないサーボモータであり，センサはロボットの姿勢クォータニオンしかセンシングできない．
  そのような複雑で不自由な身体構造において，歩行動作と起き上がり動作について，
  実機を使った試行錯誤によって発見した歩容に必要な姿勢を線形補完で接続する手法と，
  シミュレータ内での強化学習により生成した歩容を実機に反映する手法の2種類で歩容を生成し，両者を比較した．
  その結果，両者とも歩行動作および起き上がり動作に成功している一方で，強化学習によって学習されたモデルによる歩容において，
  本脚を広げながら歩く転倒リスクの低い歩行動作や，さまざまな姿勢からの起き上がり動作などの，身体構造を潜在的に考慮していると考える歩容が生成された．
}%
\end{abstract}

\section{Introduction}\label{sec:introduction}
\switchlanguage%
{%
  The movie ``Suzume''\cite{suzume2022shinkai}  by Makoto Shinkai was released on November 11, 2022.  
  In the beginning of the story, a young man is enchanted into a three-legged chair.
  The chair is a three-legged chair with one leg missing from the original four legs, resulting in an asymmetrical and incomplete leg arrangement to the body.
  The legs have only two-rotational joints at the root, and there are no joints such as elbows or knees in animals.
  The enchanted young man in the form of a chair could not move well at first, but as the story progressed, his gait grew.
  Eventually, the young man had acquired a very flexible gait, running down the street and stand-up after falling down, while still in the body structure of a chair.
  We design this interesting body structure as a chair-type asymmetric tripedal low-rigidity robot and generate its gait.

  The robot in this study is distinctive in that it has an incomplete and asymmetric body structure,
  and since it has lost one leg from its originally four-legged body, it is difficult for the robot to maintain balance.
  Also, the actuators consist of servo motors that can only provide feed-forward angular commands with severe backlash, 
  and only the roll, pitch, and yaw angles of the robot can be perceived, making the whole price of this robot cheepper to around \$60.

  We analyze what kind of gait is generated in the robot with such a body structure by generating gait using two methods and comparing them.
  one is a method that generates several necessary postures by linear interpolation,
  and the other is a method that generates gait by reinforcement learning.
  Through a comparative analysis of these two methods, this study explores gait generation in the robot with imperfect in terms of body structures and control mechanisms.


  \begin{figure}[t]
  	\centering
  	\includegraphics[width=0.8\columnwidth]{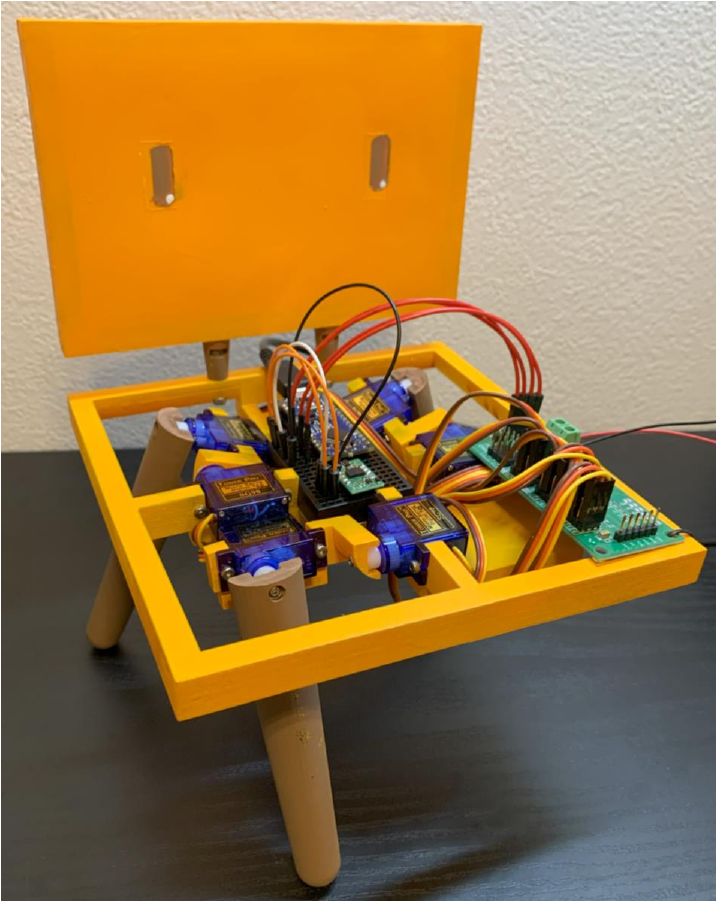}
  	\caption{Overall view of the chair-type asymmetric tripedal low-rigidity robot.}
  	\label{figure:chair_overview}
    \end{figure}
}%
{%
  映画『すずめの戸締まり』\cite{suzume2022shinkai}が2022年11月11日に公開された．
  物語の序盤で，主人公と共に旅に出る青年が姿を椅子に変えられてしまう．
  その椅子はもともとの四本脚から一本の脚が欠けた三本脚の椅子であるため体に対して非対称で不完全な脚の配置になっており，
  その脚には根元にボールジョイントのような対偶のみが存在し，動物における肘や膝などの関節は存在しない．
  その青年はそのような椅子の姿に変えられ，最初はうまく動くことができなかったものの，物語が進むにつれてその歩容は成長していき，
  最終的には椅子の身体構造のまま道を駆けたり転んでも起き上がったりと，非常に自由自在で多彩な歩容を会得していた．
  物語中の青年のセリフの中にも「だんだん体が椅子に馴染んできた」とあり，青年が椅子の姿のままその歩容を発展させていった様子が伺える．
  本研究では，これを椅子型非対称三脚移動ロボットとして再現し，歩容を生成する．
  

  %

  すでに基本的な動作が生成可能なロボットに対してより運動能力を向上させるものとして，
  外乱を受けて転倒しても起き上がって歩行を続ける四脚移動ロボット制御の研究\cite{lee2019robust}，
  四脚移動ロボットがゴールキーパーとしての動きを学習するもの\cite{huang2022creating}などがあり，
  運動能力を様々な環境に適応させるものとして，四脚移動ロボットの歩容を雪道や砂利道などの様々な環境に適応させる研究\cite{lee2020learning}がある．
  また，ロボットに戦略的なスキルを学習させるものとして，二足歩行ロボットが高度な対敵サッカースキルを学習するもの\cite{haarnoja2023learning}がある．
   
  本研究で扱うロボットは身体構造が非対称，不完全で，アクチュエータもフィードフォワードに角度指令しかできないサーボモータであり，センシング可能なものは
  ロボットのロール，ピッチ，ヨー角のみである．
  試行錯誤で発見したいくつかの姿勢を線形に遷移させる手法と，強化学習による手法とで生成し，両者を比較することで，
  身体構造や制御の観点から不自由であるロボットにおける歩容生成を考察する．
}%

\section{Related Works}\label{sec:related_works}
\subsection{Conventional control for legged robots}
\switchlanguage%
{%
  In \cite{liu2011cpg}, the central pattern generator (CPG) was used to generate leg trajectories to make the quadruped robot walk on various uneven terrain, 
  and in \cite{tanikawa2021reciprocal}, the reflex circuit for walking, which is also found in cats, was implemented in the robot to achieve steady walking, 
  but it is difficult to adapt these to the tripedal robot with an asymmetrical body structure that balances and walks.
  In \cite{bledt2018cheetah , di2018dynamic}, the quadruped robot is controlled by model predictive control (MPC) and shows high locomotion performance in various gaits, 
  such as walking on stairs and uneven terrain, running at a maximum of 3 m/s, and walking on 3 legs. 
  The actuators used in the quadruped robot are brushless motors, which provide excellent controllability and backdrivability, but
  the actuators used in this study are low-cost servo motors with poor controllability that is unable to provide torque conrol, making model-based control difficult.
}%
{%
  \cite{bledt2018cheetah, di2018dynamic}ではモデル予測制御（MPC）によって四脚ロボットが制御され，階段や不整地での歩行や
  最大3m/sの走行に加えて，三本脚での歩行など，様々な歩容において高い運動性能を誇っている．
  ロボットモデルは１つの脚につき3自由度の計12自由度で表現され，使用されているアクチュエータはブラシレスモータであり，
  バックドライバビリティに優れ，制御器が計算した通りの高トルクが実現できる．
  これらの高い運動性能を実現するのは，各脚の十分な自由度と，十分な力を発揮できる制御性の高いアクチュエータが
  あってこそである．
  本研究のロボットは，1つの脚につき2自由度の計6自由度しかなく，人間や動物が本来持つ膝関節などに当たるものが存在しない．
  また，アクチュエータは電子工作レベルのサーボモータであり， そのトルクを制御することはできず，制御できるのは位置だけであり，モデルベースに制御することは難しい．
}%
\subsection{Reinforcement learning methods for legged robots}
\switchlanguage%
{%
  In the study of quadruped robot gait generation, 
  \cite{lee2019robust} generated a gait that got up and continued walking even after falling down due to external disturbances, 
  \cite{lee2020learning} generated a gait that adapted to various environments such as snow and gravel paths, 
  and \cite{huang2022creating} learned to move as a goalkeeper.
  Also bipedal robots have acquired advanced soccer skills to compete against opponents in \cite{haarnoja2023learning}.
  Compared to these studies, the robot in this study has an incomplete and asymmetrical body structure, 
  is small (156 mm $\times$ 156 mm), and has few sensors, sensing only roll, pitch, and yaw angles.
  In addition, sim2real with low-cost servo motors that cost only about \$3 is required, 
  and the output of reinforcement learning is the command position of the servo motors themselves.
}%
{%
  外乱を受けて転倒しても起き上がって歩行を続ける研究\cite{lee2019robust}や
  雪道や砂利道などの様々な環境に歩用を適応させる研究\cite{lee2020learning}，
  ゴールキーパーとしての動きを学習するもの\cite{huang2022creating}があり，
  また，二足歩行ロボットにおいては， 高度な対敵サッカースキルを学習するもの\cite{haarnoja2023learning}がある．
  これらの研究と比較してこの研究のロボットは不完全で非対称の身体構造を持ち、その大きさは156mmかける156mmサイズに収まるような小さく、
  持っているセンサが少なくロールピッチヨー角しかセンシングすることができない。
  また，貧弱なサーボモータでのsim2realが必要であり，
  強化学習が出力するのは脚の軌跡などではなく，そのサーボモータの指令位置そのものである．
}%
\subsection{Robots with characteristic body structure}
\switchlanguage%
{%
  In \cite{kawaharazuka2022realization}, a tendon-driven musculoskeletal humanoid performs movements while seated on a chair with casters, 
  with human instructions guiding gait generation. 
  \cite{sims1994evolving} explores the generation and evolution of complex body structures and corresponding gaits within a simulator, 
  utilizing a genetic algorithm to achieve designs beyond human conception. 
  Additionally, \cite{ha2018automated} generated gate for each body structure from one to three legs on an actual robot by reinforcement learning.
  \cite{maekawa2018improvised} used a body structure of connecting trees in nature with several joints, and learned an efficient gait to go far away by reinforcement learning.
  \cite{bongard2006resilient} modeled the robot itself by randomly moving its body without knowing its own body structure, and then generated a gait that suited its body.
  The robot in this study differs from these robots in that it is chair-shaped and behaves in a chair-like way in generating gait, such as keeping its seat high and horizontal, 
  and needs to keep its balance while generating gait with an imperfect and asymmetrical body structure.
}%
{%
  複雑な身体構造で歩容を生成する前例として，姿勢を人間が教示することで腱駆動筋骨格ヒューマノイドがキャスター付きの椅子に座りながら動作する研究\cite{kawaharazuka2022realization}がある．
  遺伝的アルゴリズムを用いて，シミュレータの中で身体構造とその制御系を生成し発展させる研究\cite{sims1994evolving}では，
  人間の設計によって発明され難い身体構造と制御則を発現させた．
  また，1脚から3脚までのそれぞれの身体構造で実機ロボットでの強化学習により歩容を生成させる研究\cite{ha2018automated}もある，
  \cite{maekawa2018improvised}では，自然にある木同士をいくつかの関節でつなぐという身体構造で，強化学習により遠くへ行くための効率的な歩行を学習した．
  \cite{bongard2006resilient}では，自分の身体構造が不明な状態で体をランダムに動かし，自己をモデリングした上で，その身体にあった歩用を生成した．
  本研究のロボットは椅子型をなしており、歩容生成についてはその座面を高く水平に保つなどの椅子らしい振る舞いをするため、
  不完全で非対称な身体構造を持ちながらバランスを保ちながら歩容を生成する必要があるという点で，これらのロボットとは異なる．
}%

\section{Body Design of Chair-Shaped Asymmetrical Tripedal Robot} \label{sec:design}
\subsection{Design of Gimbal-Type 2-DOF Leg}
\switchlanguage%
{%
  \begin{figure}[t]
    \centering
    \includegraphics[width=0.7\columnwidth]{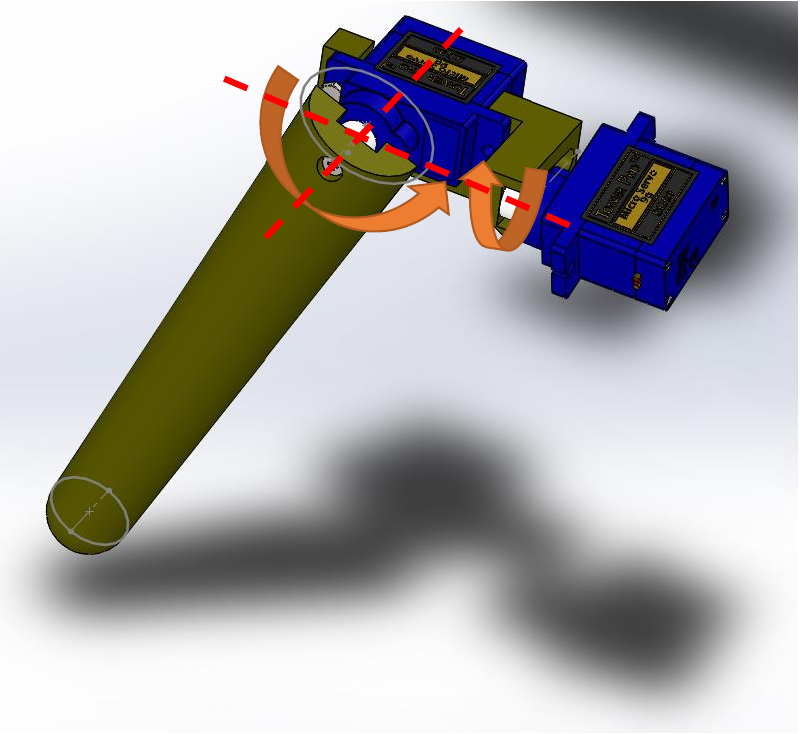}
    \caption{Gimbal-type 2-DOF leg.}
    \label{figure:leg}
    \end{figure}

  The gimbal-type 2-DOF leg designed in this study is shown in \figref{figure:leg}.
  The rotation axes of the two servo motors intersect at the apex of the leg, allowing the leg to move forward, backward, left, and right.
  The servo motor that rotates the leg is fixed to a bracket, which is rotated by another servo motor.
  Implemented fillets on the bracket ensure each servo motor has a motion range of $\pm\ang{50}$. 
  The SG-90 Micro Servo was selected, which is lightweight, space-saving, and can be driven from an Arduino.
}%
{%
  \begin{figure}[t]
  	\centering
  	\includegraphics[width=0.7\columnwidth]{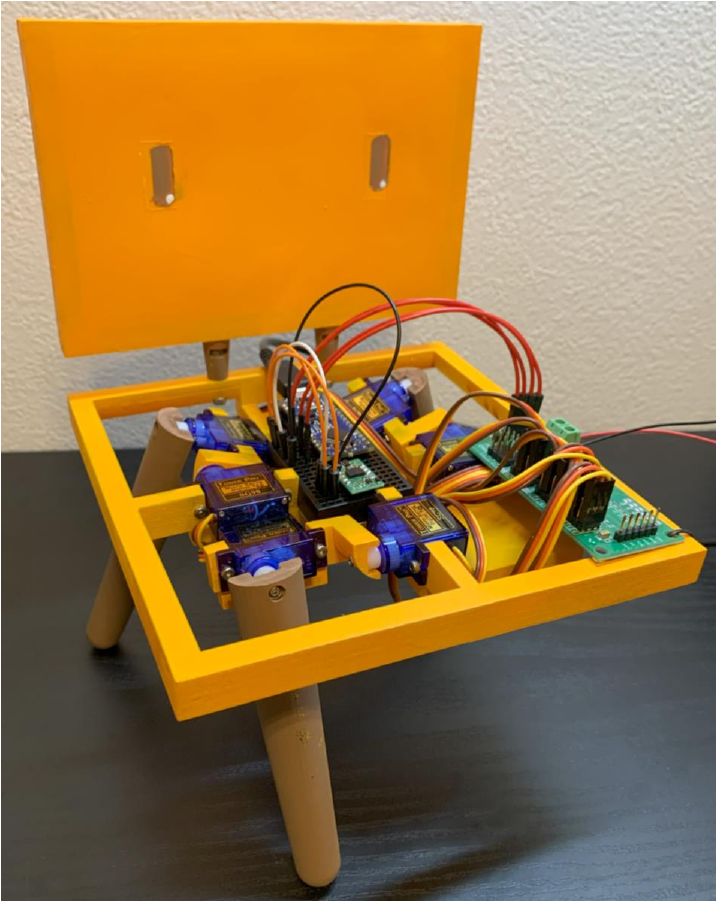}
  	\caption{椅子型非対称三脚移動ロボットの全体図.}
  	\label{figure:chair_overview}
    \end{figure}
  
  \begin{figure}[h]
    \centering
    \includegraphics[width=0.7\columnwidth]{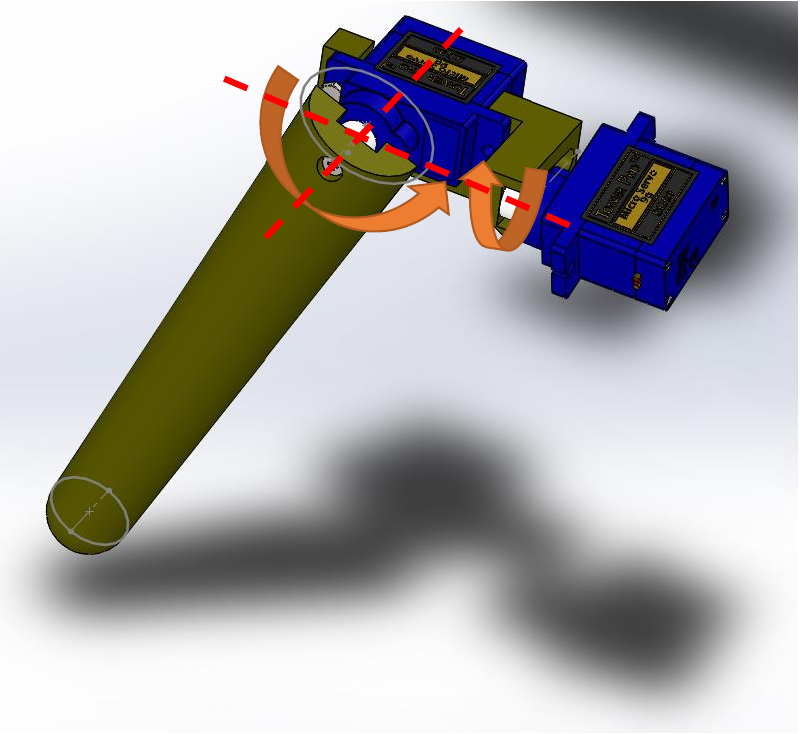}
    \caption{シンバル型2自由度脚.}
    \label{figure:leg}
    \end{figure}

  本研究で設計したジンバル型2自由度脚を\figref{figure:leg}に示す．
  2つのサーボモータの回転軸の交点が脚の頂点で交わることにより，前後左右に脚を動かすことができる．
  脚を回転させるサーボモータをブラケットに固定し，そのブラケットをもう1つのサーボモータで回転させる構造となっている．
  ブラケットに適切なフィレットをかけることにより，それぞれのサーボモータは$\pm\ang{50}$の可動域を持つ．
  サーボモータには軽量，省スペースでありArduinoから駆動できるSG-90マイクロサーボを採用した．
}%

\subsection{Overall Design of the Robot}
\switchlanguage%
{%
  The entire chair-type robot is designed by constructing three gimbal-type 2-DOF legs shown in \figref{figure:leg}.
  These legs are arranged to allow for six servo motors to fit within the plane of the chair's seat surface, 
  with the ground contact points of the legs positioned near the corners of the seat rectangle. 
  The legs' range of motion is maintained by filleting the inside of the seat surface as well as the brackets.
  In addition, the empty space created by the lack of one leg has a surface where a servo motor control board or microcontroller can be placed.
  The backrest is designed with a face similar to that of the character in the movie.
  The solid backrest would be difficult to stand on three legs due to the increased weight and bias, 
  so only the frame of the backrest was designed, and a painted clear file was pasted to create the surface.
  Based on the above, the overall view of the chair-type asymmetric tripedal low-rigidity robot designed in this study,
  fabricated with a 3D printer, and painted is shown in \figref{figure:chair_overview}.
}%
{%
  \figref{figure:leg}に示したジンバル型2自由度脚を3脚構成することで椅子型ロボット全体を設計する．
  合計6つのサーボモータが椅子の座面平面に収まりつつ，脚の接地点が座面四角形の角付近になるように3脚を配置した．
  座面の内側にもブラケット同様にフィレットをかけることで脚の可動範囲を保っている．
  また，1脚欠けていることで生まれた空きスペースにはサーボモータ制御基板やマイコンを設置できるような面を設けた．
  背もたれ部分は映画のキャラクターと同じように顔を設計した．
  ここで，中実の背もたれだと重量の増加と偏りが発生し3脚での自立が困難になったので，背もたれの枠のみを設計し，塗装したクリアファイルを貼り付けて面を作っている．
  以上の事を踏まえ本研究で設計し，3Dプリンタで造形した後に塗装した椅子型非対称三脚移動ロボットの全体図を\figref{figure:chair_overview}に示す．
}%

\subsection{System Configuration}
\switchlanguage%
{%
  The system configuration diagram of the robot in this study is shown in \figref{figure:system}.
  The robot features an Arduino Nano Every that transmits command angles to six servo motors through the servo motor control board 
  while receiving quaternions of its posture from the IMU.
  The servo motors are powered by the power supply unit via the servo motor control board.
  The Arduino Nano Every is wired for connection with an external PC, using ROS to exchange command angles and quaternions.
  This communication cycle, involving ROS, servo motors, and IMU sensor values, operates at a frequency of $\SI{10}{\hertz}$.
  The system configuration is such that the PC executes posture planning while receiving sensor values as feedback, 
  and the six command angles are sent to the Arduino Nano Every to realize walking and stand-up motions.

  \begin{figure}[t]
  	\centering
  	\includegraphics[width=1\columnwidth]{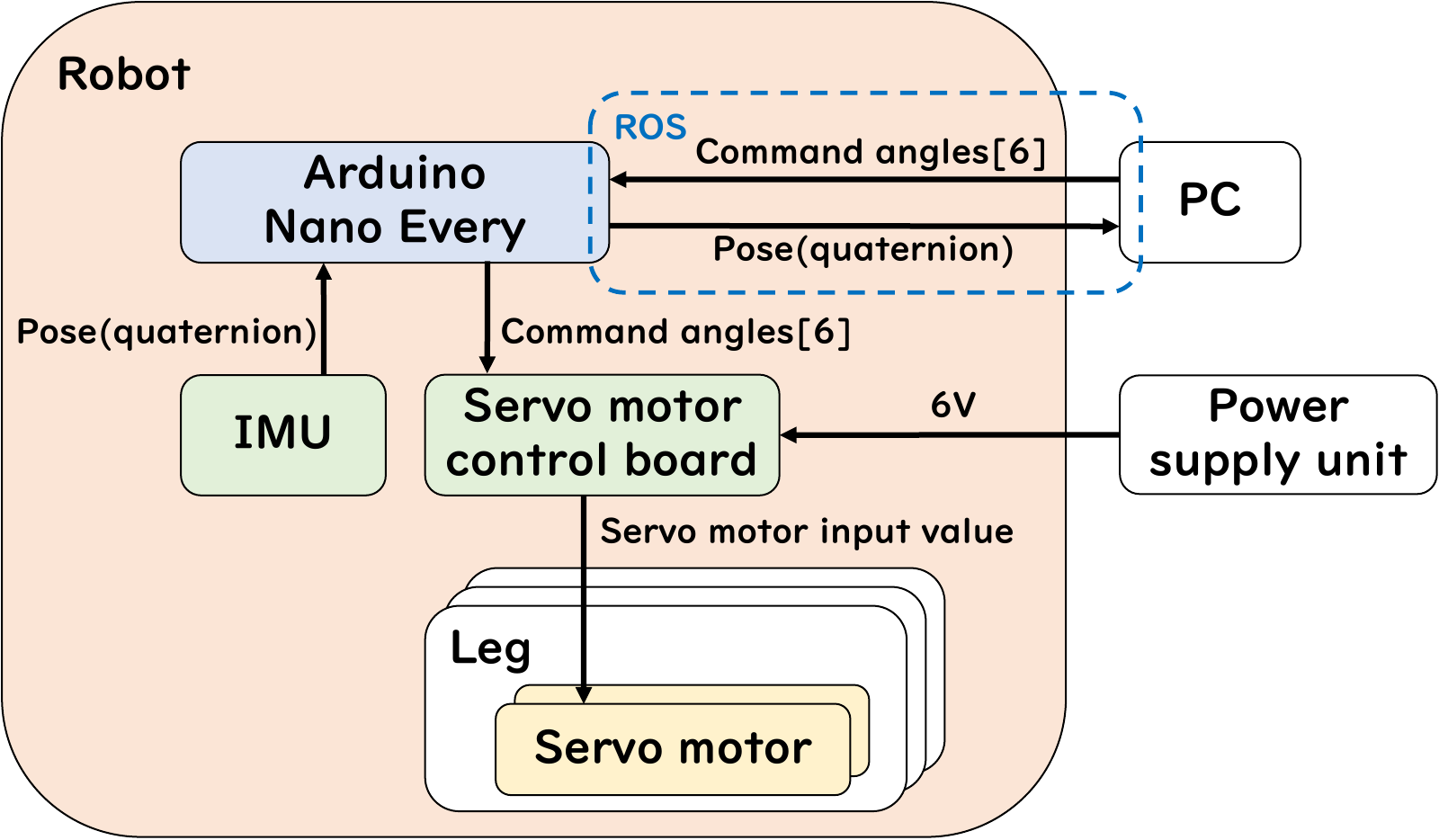}
  	\caption{System configuration.}
  	\label{figure:system}
    \end{figure}
}%
{%
  本研究のロボットのシステム構成図を\figref{figure:system}に示す．
  ロボットにはArduino Nano Everyが搭載されており，サーボモータ制御基板を介してサーボモータ6つに指令角度を送信し，IMUからロボットの姿勢のクォータニオンを受信する．
  サーボモータの駆動電源はサーボモータ制御基板を介して電源装置から供給される．
  Arduino Nano Everyは外部のPCと有線で接続され，ROSを用いることで指令角度とクォータニオンを通信する．
  ROS，サーボモータ，IMUセンサ値の通信周期は$\SI{10}{\hertz}$で行う．
  これにより，PC上でセンサ値をフィードバックとして受けながら姿勢計画を実行し，
  その結果として6つの指令角度をArduino Nano Everyに送ることで歩行や起き上がりの動作を実現するというシステム構成になっている．

  \begin{figure}[t]
  	\centering
  	\includegraphics[width=1\columnwidth]{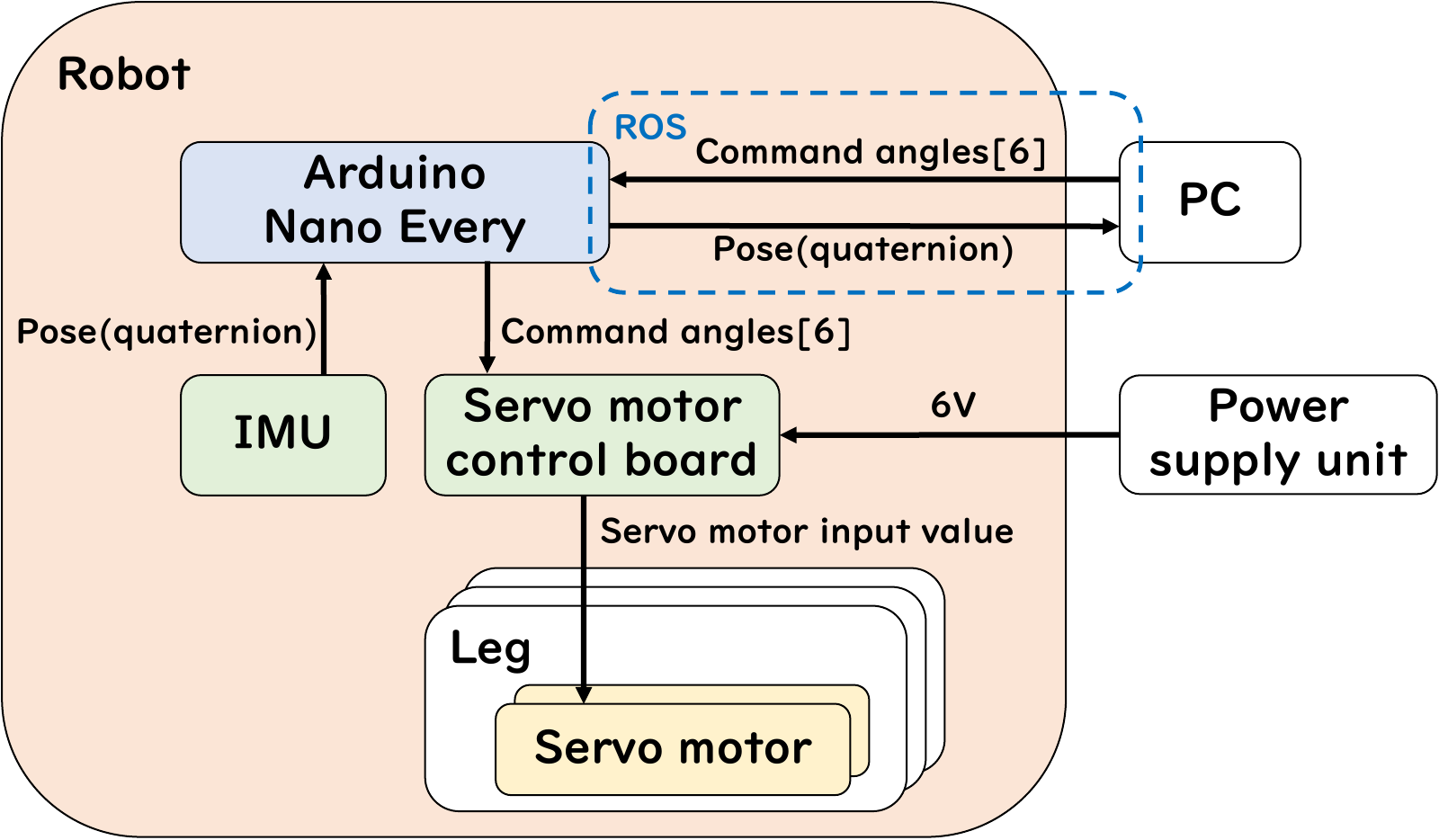}
  	\caption{システム構成図.}
  	\label{figure:system}
    \end{figure}
}%

\section{Gait Generation of Chair-Shaped Asymmetrical Tripedal Robot}
\switchlanguage%
{%
  In this study, two methods were employed to generate the gait of a chair-type asymmetric tripedal low-rigidity robot. 
  The first method involves gait generation through connecting essential postures, while the second method involves gait generation through reinforcement learning. 
  This chapter provides a detailed description of the steps and methods employed for each gait generation approach.
}%
{%
  本研究では，2つの方法により椅子型非対称三脚移動ロボットの歩容を生成した．
  1つ目に姿勢遷移による発見的な歩容生成，2つ目に強化学習による歩容生成である．
  この章では，それぞれの歩容生成について具体的な手順と方法を説明する．
}%

\subsection{Gait Generation through Connecting Essential Posture} 
\switchlanguage%
{%
  The posture of the robot is determined by the angles of its six servo motors, which are then represented as a single point in a six-dimensional space.
  To generate gait, the essential postures required for the motion process are discovered through experimentation with the actual robot.
  The robot's gait is generated by transitioning between these postures using linear interpolation.
  A posture $S$ denotes a 6-dimensional vector containing the angles of the six servo motors.
  $$\bm{S} = \begin{bmatrix} \theta_0 & \theta_1 & \theta_2 & \theta_3 & \theta_4 & \theta_5 \end{bmatrix}^\top$$
  The process of connecting essential postures using linear interpolation can be expressed as follows: 
  when moving from one posture $\bm{S}_i$ to the subsequent posture $\bm{S}_{i+1}$, the interpolation posture $\bm{s}$ is employed.
  \begin{equation*}
    \begin{split}
      \begin{bmatrix}\bm{S}_i & \bm{S}_{i+1} \end{bmatrix} \begin{bmatrix} 1 & \frac{{n}}{{n+1}} & \frac{{n-1}}{{n+1}} & \ldots & \frac{{1}}{{n+1}} & 0 \\ 0 & \frac{1}{n+1} & \frac{2}{n+1} & \ldots & \frac{n}{n+1} & 1\end{bmatrix} 
        \\= \begin{bmatrix}\bm{S}_i & \bm{s}_{1} & \bm{s}_{2} & \ldots & \bm{s}_{n} &  \bm{S}_{i+1} \end{bmatrix}
    \end{split}
  \end{equation*}

  Using these expressions, the following concrete steps are taken
  \begin{enumerate}
		\item Determine the initial posture $\bm{S}_0$.
		\item Determine the end posture $\bm{S}_f$.
		\item Tentatively determine the essential posture $\bm{S}_i$ manually in the connecting essential postures process.
		\item Transition and operate the actual robot posture from $\bm{S}_0$ to $\bm{S}_f$ via $\bm{S}_i$ by linear interpolation in $n_i$ steps.
		\item Modify $n_i$, or modify / add / delete $\bm{S}_i$ so that the motion is achieved.
		\item Repeat (4) and (5) until the motion is achieved.
  \end{enumerate}

  $\theta_j \quad (j=0,1,2,3,4,5)$ corresponds to each servo motor as in \figref{figure:leg_theta}.
  $\theta_j=\ang{90}$ in the posture where all three legs are vertical to the ground.

  \begin{figure}[htb]
  	\centering
  	\includegraphics[width=0.7\columnwidth]{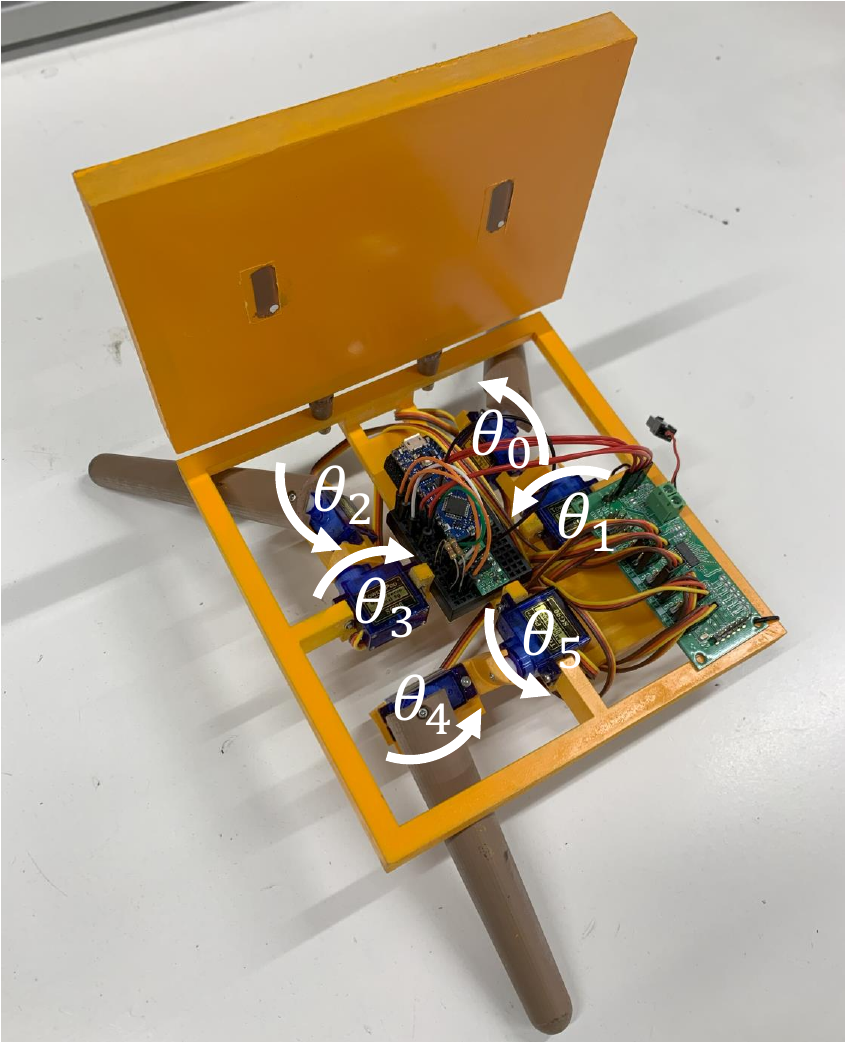}
  	\caption{Correspondence between the servo motors and $\theta_j$.}
  	\label{figure:leg_theta}
    \end{figure}
}%
{%
  ロボットの姿勢はサーボモータ6つ分の角度で決定されるので，6次元空間のある1点で表現できる．
  姿勢遷移による発見的な歩容生成では，動作の過程に必要な姿勢を実機を用いた試行錯誤により発見し，それらの姿勢を線形補間で辿って遷移させることで歩容を生成する．
  ここで，姿勢$S$はサーボモータ6つ分の角度を持つ6次元縦ベクトルである．
  $$\bm{S} = \begin{bmatrix} \theta_0 & \theta_1 & \theta_2 & \theta_3 & \theta_4 & \theta_5 \end{bmatrix}^\top$$
  線形補間による姿勢遷移について，ある姿勢$\bm{S}_i$から$\bm{S}_{i+1}$へ$n$ステップの線形補間姿勢$\bm{s}$で遷移する場合，以下のように表現する．
  \begin{equation*}
    \begin{split}
      \begin{bmatrix}\bm{S}_i & \bm{S}_{i+1} \end{bmatrix} \begin{bmatrix} 1 & \frac{{n}}{{n+1}} & \frac{{n-1}}{{n+1}} & \ldots & \frac{{1}}{{n+1}} & 0 \\ 0 & \frac{1}{n+1} & \frac{2}{n+1} & \ldots & \frac{n}{n+1} & 1\end{bmatrix} 
        \\= \begin{bmatrix}\bm{S}_i & \bm{s}_{1} & \bm{s}_{2} & \ldots & \bm{s}_{n} &  \bm{S}_{i+1} \end{bmatrix}
    \end{split}
  \end{equation*}

  これらを用いて，具体的には次の手順をたどる．
  \begin{enumerate}
    \item 初期姿勢$\bm{S}_0$を決定する
    \item 終了姿勢$\bm{S}_f$を決定する
    \item 姿勢遷移過程における姿勢$\bm{S}_i \quad (i=1, 2, 3,\ldots)$を仮決定する
    \item ロボット実機の姿勢を$\bm{S}_0$から$\bm{S}_i$を介して$\bm{S}_f$まで$n_i$ステップで線形補間して遷移させ動作させる
    \item 動作が達成されるように$n_i$を修正，$\bm{S}_i$を修正，追加，削除する
    \item 動作が達成されるまで(4), (5)を繰り返す
  \end{enumerate}
  
  $\theta_j \quad (j=0,1,2,3,4,5)$はそれぞれのサーボモータに\figref{figure:leg_theta}のように対応する．
    また，3脚すべてを地面に対して垂直にした姿勢において$\theta_j=\ang{90}$である．
  
  \begin{figure}[htb]
  	\centering
  	\includegraphics[width=0.5\columnwidth]{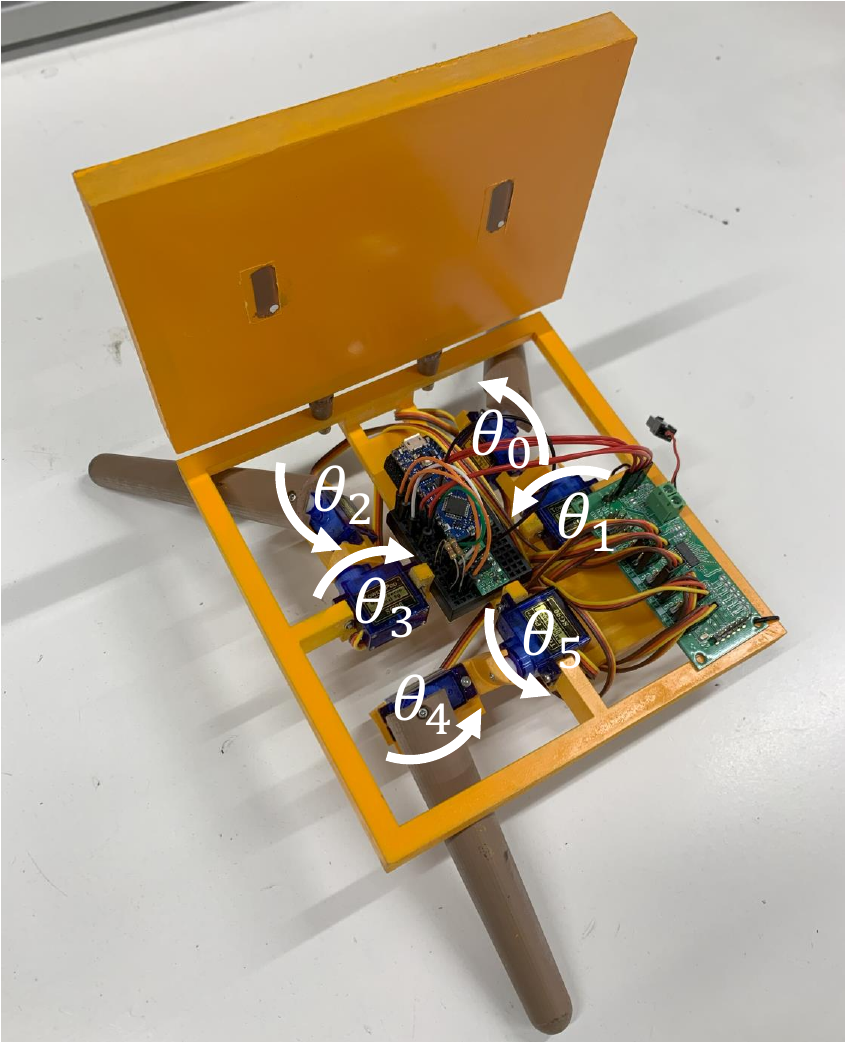}
  	\caption{サーボモータと$\theta_j$との対応.}
  	\label{figure:leg_theta}
    \end{figure}
}%

\subsubsection{Walking Motion}
\switchlanguage%
{%
  \begin{figure}[htb]
  	\centering
  	\includegraphics[width=1\columnwidth]{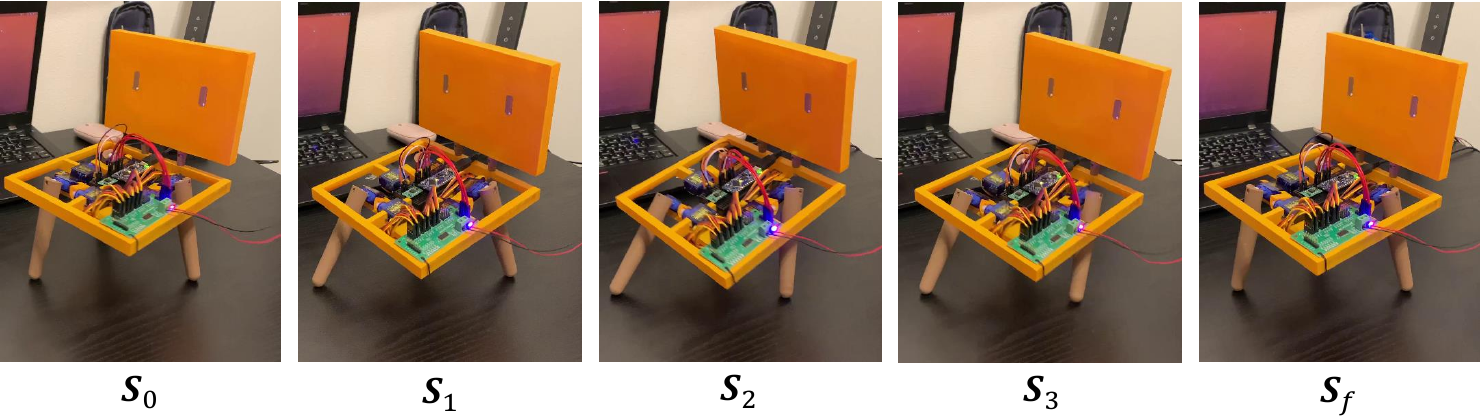}
  	\caption{Walking motion by connectiong essential posture.}
  	\label{figure:walk_trans}
    \end{figure}

  \begin{table}[htb]
  	\centering
    \caption{Posture and the number of linear interpolations in walking motion.}
  	\label{table:walk_table}
    \begin{tabular}{|c||l|l|} \hline
       & Posture & Steps \\ \hline \hline
      $\bm{S}_0$ & $\begin{bmatrix} \ang{90} & \ang{80} & \ang{90} & \ang{100} & \ang{90} & \ang{100} \end{bmatrix}^\top$ & \multirow{2}{*}{0} \\
      $\bm{S}_1$ & $\begin{bmatrix} \ang{90} & \ang{80} & \ang{90} & \ang{100} & \ang{90} & \ang{140} \end{bmatrix}^\top$ & \multirow{2}{*}{0} \\
      $\bm{S}_2$ & $\begin{bmatrix} \ang{70} & \ang{80} & \ang{90} & \ang{100} & \ang{90} & \ang{140} \end{bmatrix}^\top$ & \multirow{2}{*}{0} \\
      $\bm{S}_3$ & $\begin{bmatrix} \ang{70} & \ang{80} & \ang{110} & \ang{100} & \ang{90} & \ang{140} \end{bmatrix}^\top$ & \multirow{2}{*}{6} \\
      $\bm{S}_f$ & $\begin{bmatrix} \ang{90} & \ang{80} & \ang{90} & \ang{100} & \ang{90} & \ang{100} \end{bmatrix}^\top$ &   \\ \hline
      \end{tabular}
    \end{table}

  The postures and the number of linear interpolations found in the walking motion are shown in \figref{figure:walk_trans} and \tabref{table:walk_table}.
  This walking motion sets a standing posture for the initial and final postures, and puts the front foot forward at $\bm{S}_1$, 
  the left rear foot forward at $\bm{S}_2$, and the right rear foot forward at $\bm{S}_3$. 
  By repeating these posture transitions, the robot walks.
}%
{%
  \begin{figure}[htb]
  	\centering
  	\includegraphics[width=1\columnwidth]{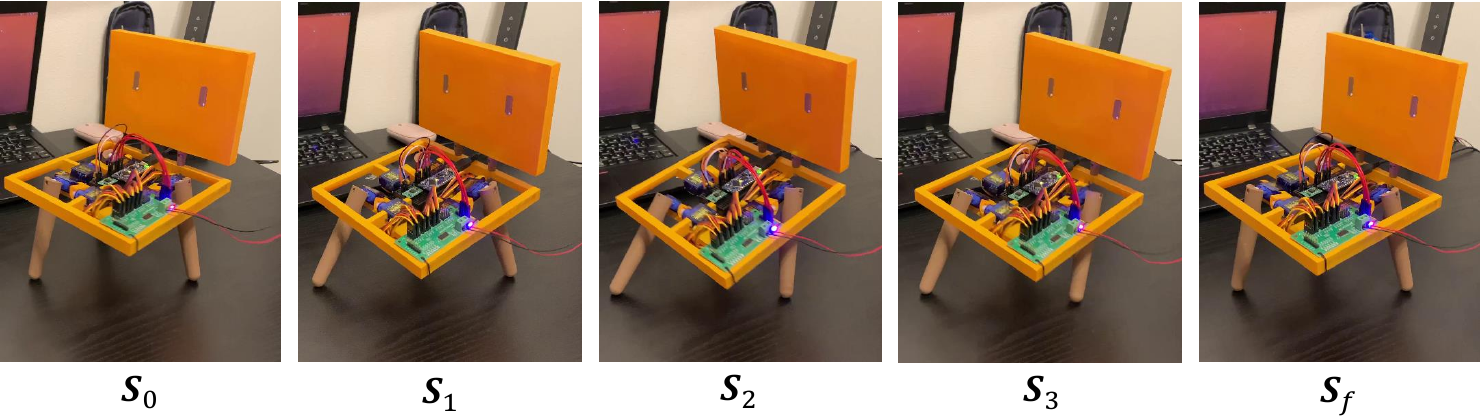}
  	\caption{姿勢遷移による歩行動作.}
  	\label{figure:walk_trans}
    \end{figure}
  
  \begin{table}[htb]
  	\centering
    \caption{歩行動作における姿勢と線形補間数.}
  	\label{table:walk_table}
    \begin{tabular}{|c||l|l|} \hline
       & 姿勢 & 補間数 \\ \hline \hline
      $\bm{S}_0$ & $\begin{bmatrix} \ang{90} & \ang{80} & \ang{90} & \ang{100} & \ang{90} & \ang{100} \end{bmatrix}^\top$ & \multirow{2}{*}{0} \\
      $\bm{S}_1$ & $\begin{bmatrix} \ang{90} & \ang{80} & \ang{90} & \ang{100} & \ang{90} & \ang{140} \end{bmatrix}^\top$ & \multirow{2}{*}{0} \\
      $\bm{S}_2$ & $\begin{bmatrix} \ang{70} & \ang{80} & \ang{90} & \ang{100} & \ang{90} & \ang{140} \end{bmatrix}^\top$ & \multirow{2}{*}{0} \\
      $\bm{S}_3$ & $\begin{bmatrix} \ang{70} & \ang{80} & \ang{110} & \ang{100} & \ang{90} & \ang{140} \end{bmatrix}^\top$ & \multirow{2}{*}{6} \\
      $\bm{S}_f$ & $\begin{bmatrix} \ang{90} & \ang{80} & \ang{90} & \ang{100} & \ang{90} & \ang{100} \end{bmatrix}^\top$ &   \\ \hline
      \end{tabular}
    \end{table}
  
  歩行動作において発見した姿勢と線形補間数を\figref{figure:walk_trans}と\tabref{table:walk_table}に示す．
  この歩行動作は初期姿勢と終了姿勢に起立姿勢を設定し，$\bm{S}_1$で前足を前に，$\bm{S}_2$で左後ろ足を前に，$\bm{S}_3$で右後ろ足を前に出す．この姿勢遷移を繰り返すことでロボットは歩行する．
}%

\subsubsection{Stand-up Motion}
\switchlanguage%
{%
  \begin{figure}[htb]
  	\centering
  	\includegraphics[width=1\columnwidth]{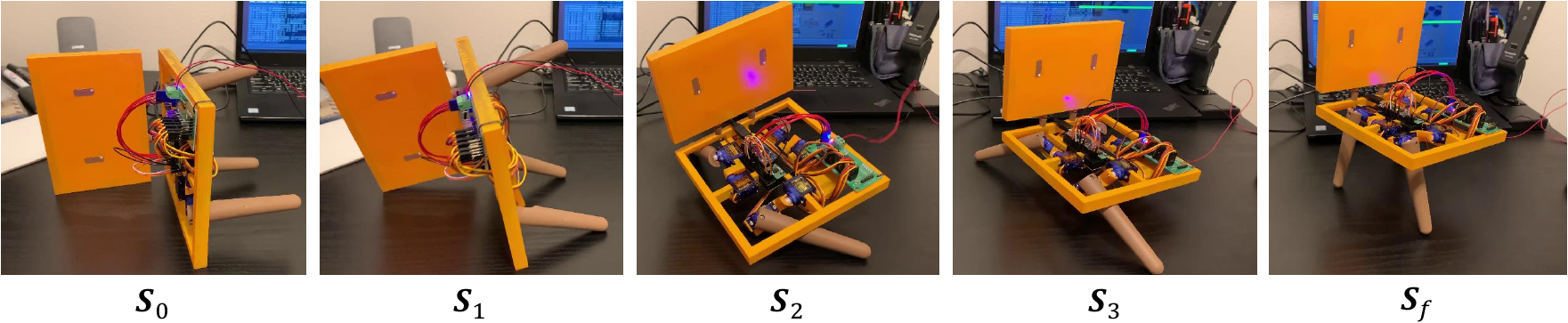}
  	\caption{Stand-up motion by posture transition.}
  	\label{figure:stand_trans}
    \end{figure}

  \begin{table}[htb]
  	\centering
    \caption{Posture and the number of linear interpolations in stand-up motion.}
  	\label{table:stand_table}
    \begin{tabular}{|c||l|l|} \hline
       & Posture & Steps \\ \hline \hline
      $\bm{S}_0$ & $\begin{bmatrix} \ang{90} & \ang{80} & \ang{90} & \ang{100} & \ang{90} & \ang{100} \end{bmatrix}^\top$ & \multirow{2}{*}{8} \\
      $\bm{S}_1$ & $\begin{bmatrix} \ang{90} & \ang{80} & \ang{90} & \ang{140} & \ang{40} & \ang{90} \end{bmatrix}^\top$ & \multirow{2}{*}{0} \\
      $\bm{S}_2$ & $\begin{bmatrix} \ang{130} & \ang{80} & \ang{40} & \ang{90} & \ang{150} & \ang{140} \end{bmatrix}^\top$ & \multirow{2}{*}{38} \\
      $\bm{S}_3$ & $\begin{bmatrix} \ang{130} & \ang{40} & \ang{50} & \ang{140} & \ang{90} & \ang{140} \end{bmatrix}^\top$ & \multirow{2}{*}{28} \\
      $\bm{S}_f$ & $\begin{bmatrix} \ang{90} & \ang{80} & \ang{90} & \ang{100} & \ang{90} & \ang{100} \end{bmatrix}^\top$ &   \\ \hline
      \end{tabular}
    \end{table}
  
  The posture and the number of linear interpolations found in the stand-up motion are shown in \figref{figure:stand_trans} and \tabref{table:stand_table}.
  In this stand-up motion, the initial posture is set with the robot's right side surface on the ground, and the end posture is set with the robot in a standing posture. 
  At $\bm{S}_1$, the robot tilts its seat by extending its legs in the opposite direction to that in which the seat is stand-up, 
  and at $\bm{S}_2$, the robot turns its seat upward by folding its legs and using their reaction.
  Then, at $\bm{S}_3$, the legs move outward to the end posture.
}%
{%
  \begin{figure}[htb]
  	\centering
  	\includegraphics[width=1\columnwidth]{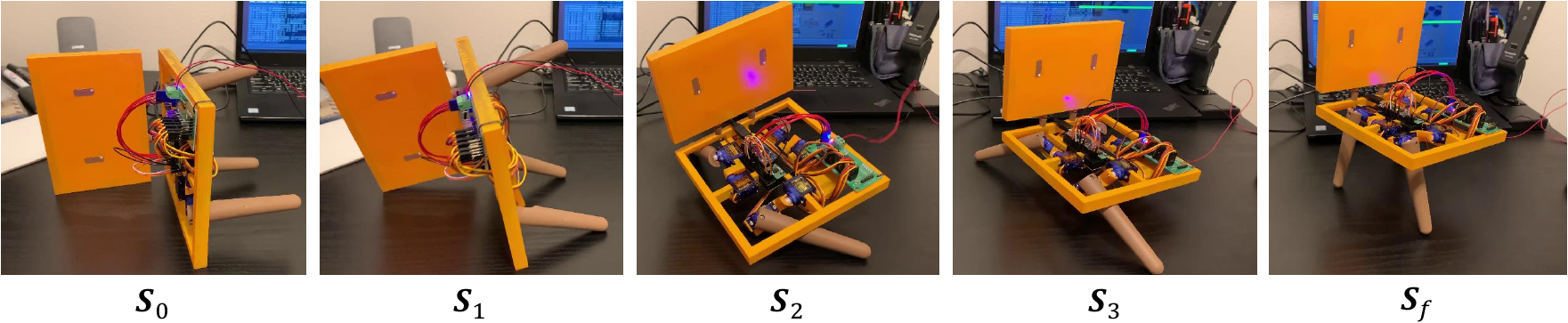}
  	\caption{姿勢遷移による起き上がり動作.}
  	\label{figure:stand_trans}
    \end{figure}
  
  \begin{table}[htb]
  	\centering
    \caption{起き上がりにおける姿勢と線形補間数.}
  	\label{table:stand_table}
    \begin{tabular}{|c||l|l|} \hline
       & 姿勢 & 補間数 \\ \hline \hline
      $\bm{S}_0$ & $\begin{bmatrix} \ang{90} & \ang{80} & \ang{90} & \ang{100} & \ang{90} & \ang{100} \end{bmatrix}^\top$ & \multirow{2}{*}{8} \\
      $\bm{S}_1$ & $\begin{bmatrix} \ang{90} & \ang{80} & \ang{90} & \ang{140} & \ang{40} & \ang{90} \end{bmatrix}^\top$ & \multirow{2}{*}{0} \\
      $\bm{S}_2$ & $\begin{bmatrix} \ang{130} & \ang{80} & \ang{40} & \ang{90} & \ang{150} & \ang{140} \end{bmatrix}^\top$ & \multirow{2}{*}{38} \\
      $\bm{S}_3$ & $\begin{bmatrix} \ang{130} & \ang{40} & \ang{50} & \ang{140} & \ang{90} & \ang{140} \end{bmatrix}^\top$ & \multirow{2}{*}{28} \\
      $\bm{S}_f$ & $\begin{bmatrix} \ang{90} & \ang{80} & \ang{90} & \ang{100} & \ang{90} & \ang{100} \end{bmatrix}^\top$ &   \\ \hline
      \end{tabular}
    \end{table}
  
  起き上がり動作において発見した姿勢と線形補間数を\figref{figure:stand_trans}と\tabref{table:stand_table}に示す．
  この起き上がり動作は，初期姿勢にロボット右側面が地面についた姿勢，終了姿勢に起立姿勢を設定し，
  $\bm{S}_1$で座面が起き上がる方向とは逆の方向に脚を伸ばすことで座面を傾け，$\bm{S}_2$で伸ばした脚をたたむことでその反動を用いて座面を上に向ける．
  そして，$\bm{S}_3$で脚を外側に出し，終了姿勢へ遷移する．
}%

\subsection{Gait Generation through Reinforcement Learning}
\switchlanguage%
{%
  The second method of gait generation is through reinforcement learning.
  In this study, walking and stand-up motions are generated by the reinforcement learning algorithm Proximal Policy Optimization (PPO) \cite{schulman2017proximal}
  in the physical simulation environment called Isaac Gym \cite{makoviychuk2021isaac}.
  The robot model used in the simulation is represented by MJCF (MuJoCo XML File), a modeling file for the MuJoCo simulator \cite{todorov2012mujoco}.
  The model includes legs, brackets, seat, backrest, and servo motor joints, as shown in \figref{figure:chair_mujoco}.
  The actuator parameters in MuJoCo were hand-tuned so that the behavior was close to reality 
  by moving the leg in the simulation while viewing the MuJoCo viewer after the theoretical values were set.

  \begin{figure}[htb]
  	\centering
  	\includegraphics[width=1\columnwidth]{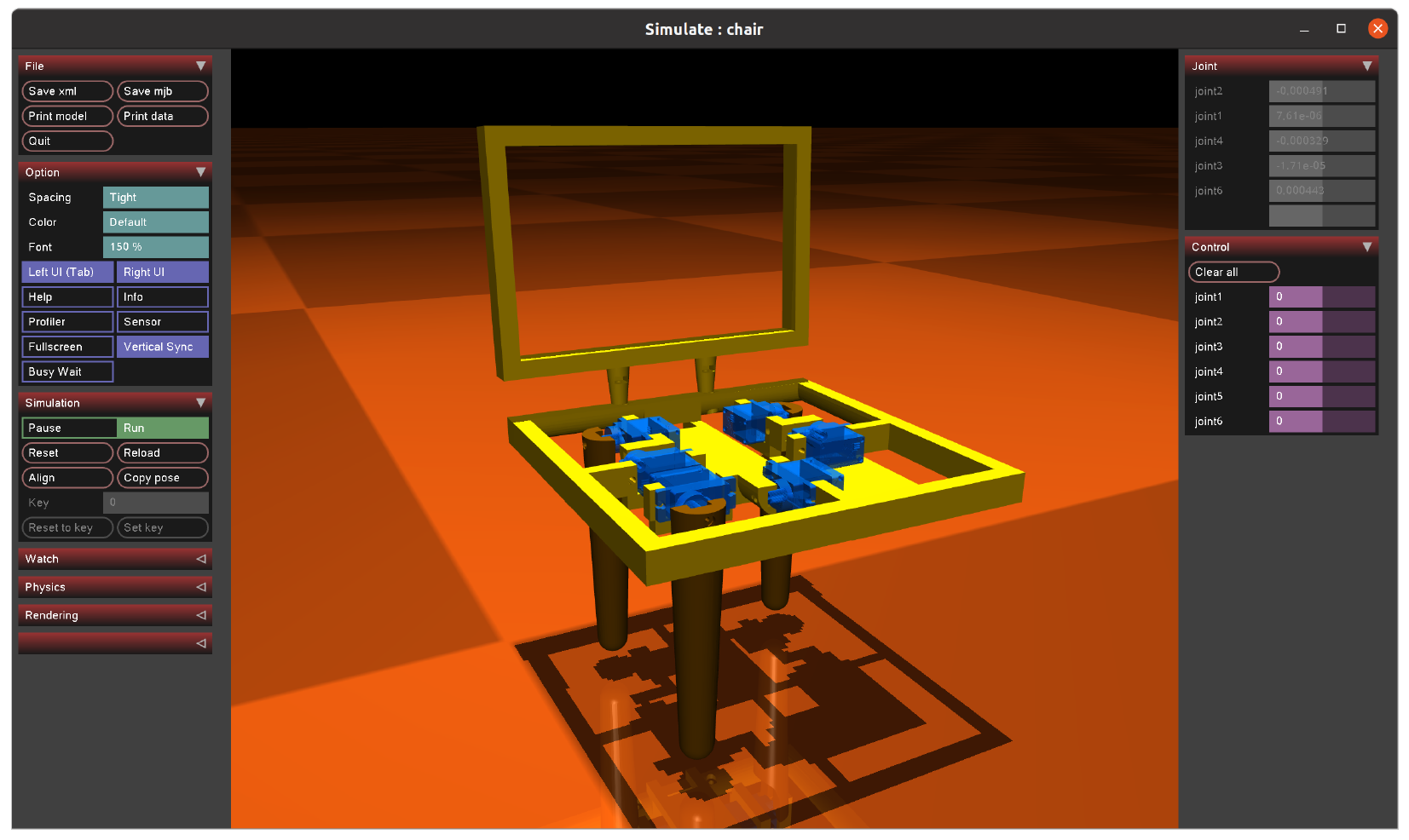}
  	\caption{Robot model used in reinforcement learning simulation.}
  	\label{figure:chair_mujoco}
    \end{figure}
  
  For reinforcement learning, the action $\bm{a} \in \mathbb{R}^6$ is the command angle for 6 servo motors, 
  and the observation $\bm{s} \in \mathbb{R}^{40}$ is the combination of the quaternion output by IMU and the servo motor command angle for four periods.
  In other words, the next action is determined from the time series data for the last 4 control cycles of the posture (quaternion) and the servo motor command angle.
  Reward and reset conditions are defined separately for walking and stand-up motions.
  The result of training is an inference model whose input is an observation $\bm{s}$ and output is an action $\bm{a}$.


  The IMU sensor values and the servo motor command angle in the actual robot are input to the inference model in the same format as during learning, 
  and the servo motor command angle, which is the next action, is obtained.
	By sending this command angle to the actual robot, what is learned in the simulation is reflected in the actual robot.

  The following variables are introduced to explain the rewards and reset conditions.
  \begin{itemize}
    \item $\bm{p} = \begin{bmatrix} x & y & z \end{bmatrix}^\top$\par : Current coordinates of the robot (center of the seat)
    \item $\bm{q} = \begin{bmatrix} q_x & q_y & q_z & q_w \end{bmatrix}^\top$\par : Posture of the seat
    \item $\bm{R_q}$\par : Rotation matrix of $\bm{q}$
    \item $\bm{p}_\text{target} = \begin{bmatrix} 10 & 0 & 0 \end{bmatrix}^\top$\par : Target coordinates / m
    \item $\text{dt} = 0.1$\par : Period of simulator / s
    \item $\bm{e}_x = \begin{bmatrix} 1 & 0 & 0 \end{bmatrix}^\top$\par : Forward unit vector
    \item $\bm{e}_z = \begin{bmatrix} 0 & 0 & 1 \end{bmatrix}^\top$\par : Vertical upward unit vector
    \item $\bm{a} = \begin{bmatrix} \theta_0 & \theta_1 & \theta_2 & \theta_3 & \theta_4 & \theta_5 \end{bmatrix}^\top$\par : Servo motor command angle
    \item $\bm{\omega} = \begin{bmatrix} \omega_0 & \omega_1 & \omega_2 & \omega_3 & \omega_4 & \omega_5 \end{bmatrix}^\top$\par : Servo motor angular velocity
    \item $u_\text{prj} = |\bm{R_q}\bm{e}_z|_z$\par : z component of the seat upward unit vector
  \end{itemize}
}%
{%
  2つ目の歩容生成の手法として強化学習による歩容生成を説明する．
  本研究では，Isaac Gym\cite{makoviychuk2021isaac}というNVIDIAの強化学習研究用の物理シミュレーション環境で歩行動作と起き上がり動作を強化学習アルゴリズムPPO\cite{schulman2017proximal}によって生成する．
  
  シミュレーション内で使われるロボットモデルはMuJoCoというシミュレータ用のモデリングファイルであるMJCF(MuJoCo XML File)により表現する．
  MJCFでは，脚，ブラケット，座面，背もたれ，サーボモータによる関節をモデリングした．MuJoCoでそれを表示したものが\figref{figure:chair_mujoco}である．
  
  \begin{figure}[htb]
  	\centering
  	\includegraphics[width=1\columnwidth]{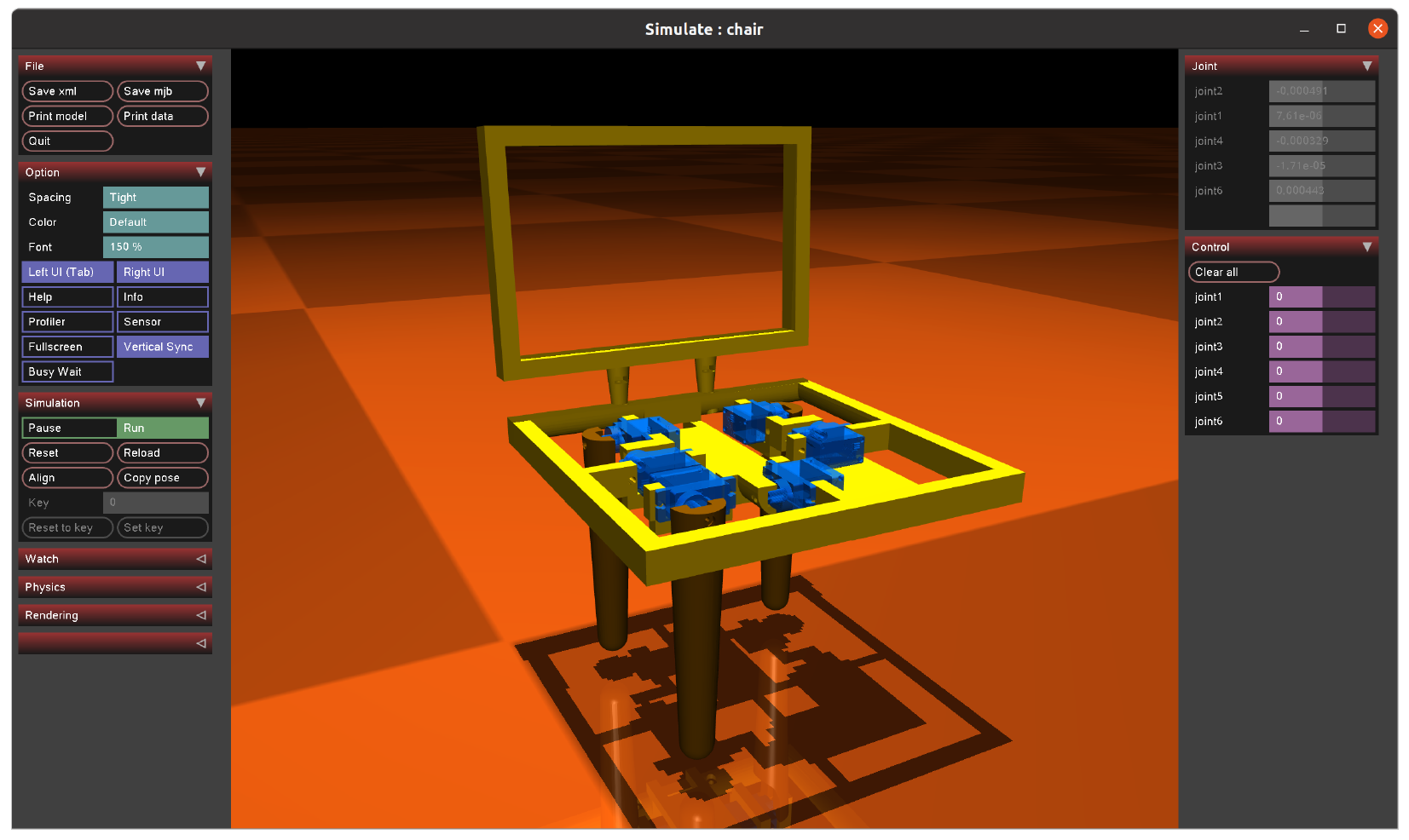}
  	\caption{強化学習シミュレーションで使用するロボットモデル.}
  	\label{figure:chair_mujoco}
    \end{figure}
  
  強化学習について，行動$a \in \mathbb{R}^6$はサーボモータ６つ分の指令角度，観測$s \in \mathbb{R}^{40}$はIMUが出力するクォータニオンとサーボモータ指令角度とを4周期分合わせたものである．
  つまり，最近\SI{200}{\milli\second}分の姿勢（クォータニオン）とサーボモータ指令角度の時系列データから次の行動を決定する．
  報酬およびリセット条件は歩行動作と起き上がり動作で別々に定義する．
  学習の結果として，入力が観測$s$，出力が行動$a$となる推論モデル\figref{figure:model}を得る．
  
  \begin{figure}[htb]
  	\centering
  	\includegraphics[width=0.4\columnwidth]{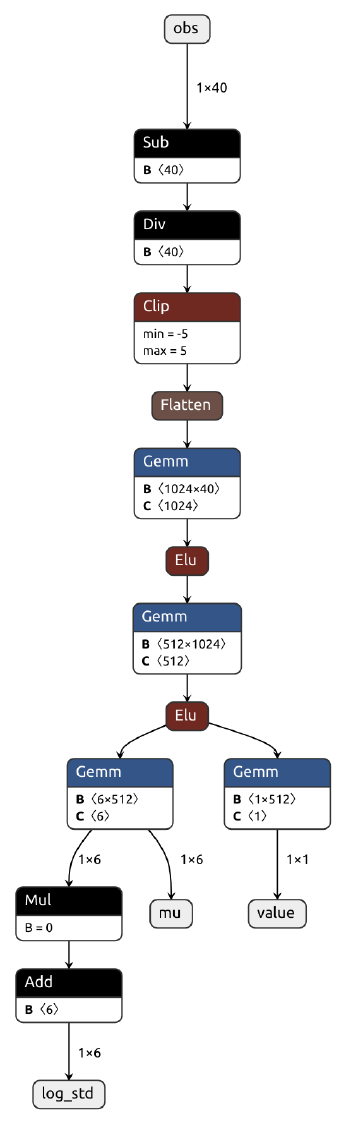}
    \caption{強化学習によって学習されるモデル.}
  	\label{figure:model}
    \end{figure}
  
  その推論モデルに対して，実機でのIMUセンサ値やサーボモータ指令角度を学習時と同じ形式に整形して入力することで，次の行動であるサーボモータ指令角度が得られる．
  これを実機に送信することで， シミュレーションで学習したものを実機に反映させる．

  報酬やリセット条件の説明のため，以下の変数を導入する．
	
  \begin{itemize}
    \item $\bm{p} = \begin{bmatrix} x & y & z \end{bmatrix}^\top$ : ロボットの現在座標(座面中心の座標)
    \item $\bm{q} = \begin{bmatrix} q_x & q_y & q_z & q_w \end{bmatrix}^\top$ : 座面の姿勢
    \item $\bm{R_q}$ : $\bm{q}$の回転行列
    \item $\bm{p}_\text{target} = \begin{bmatrix} 10 & 0 & 0 \end{bmatrix}^\top$ : 目標座標 / m
    \item $\text{dt} = 0.1$ : シュミレータの周期 / s
    \item $\bm{e}_x = \begin{bmatrix} 1 & 0 & 0 \end{bmatrix}^\top$ : 前向き単位ベクトル
    \item $\bm{e}_z = \begin{bmatrix} 0 & 0 & 1 \end{bmatrix}^\top$ : 鉛直上向き単位ベクトル
    \item $\bm{a} = \begin{bmatrix} \theta_0 & \theta_1 & \theta_2 & \theta_3 & \theta_4 & \theta_5 \end{bmatrix}^\top$ : サーボモータ指令角度
    \item $\bm{\omega} = \begin{bmatrix} \omega_0 & \omega_1 & \omega_2 & \omega_3 & \omega_4 & \omega_5 \end{bmatrix}^\top$ : サーボモータ角速度
    \item $u_\text{prj} = |\bm{R_q}\bm{e}_z|_z$ : 座面上向き単位ベクトルのz成分
  \end{itemize}
}%

\subsubsection{Walking Motion}
\switchlanguage%
{%
  The rewards and their weights are summarized in \tabref{table:walk_reward_tab} and the reset conditions are summarized in \tabref{table:walk_reset_tab}.
  Rewards are designed to be highest when moving forward while keeping the seat horizontal and high.
  However, a gait in which the robot moves forward while rubbing its back against the ground was learned as a case in which the reward was high 
  even though the robot was not in such a state. 
  The reset conditions were set to prevent such a gait from being learned.
	After training about 100 epochs of 131072 agents simultaneously, 
  about 30 epochs of additional training were conducted with noise randomly added to physical parameters, sensor values and action values.
	The walking behavior on the simulator after training is shown in \figref{figure:issacgym_walk}.

  \begin{figure}[htb]
  	\centering
  	\includegraphics[width=1\columnwidth]{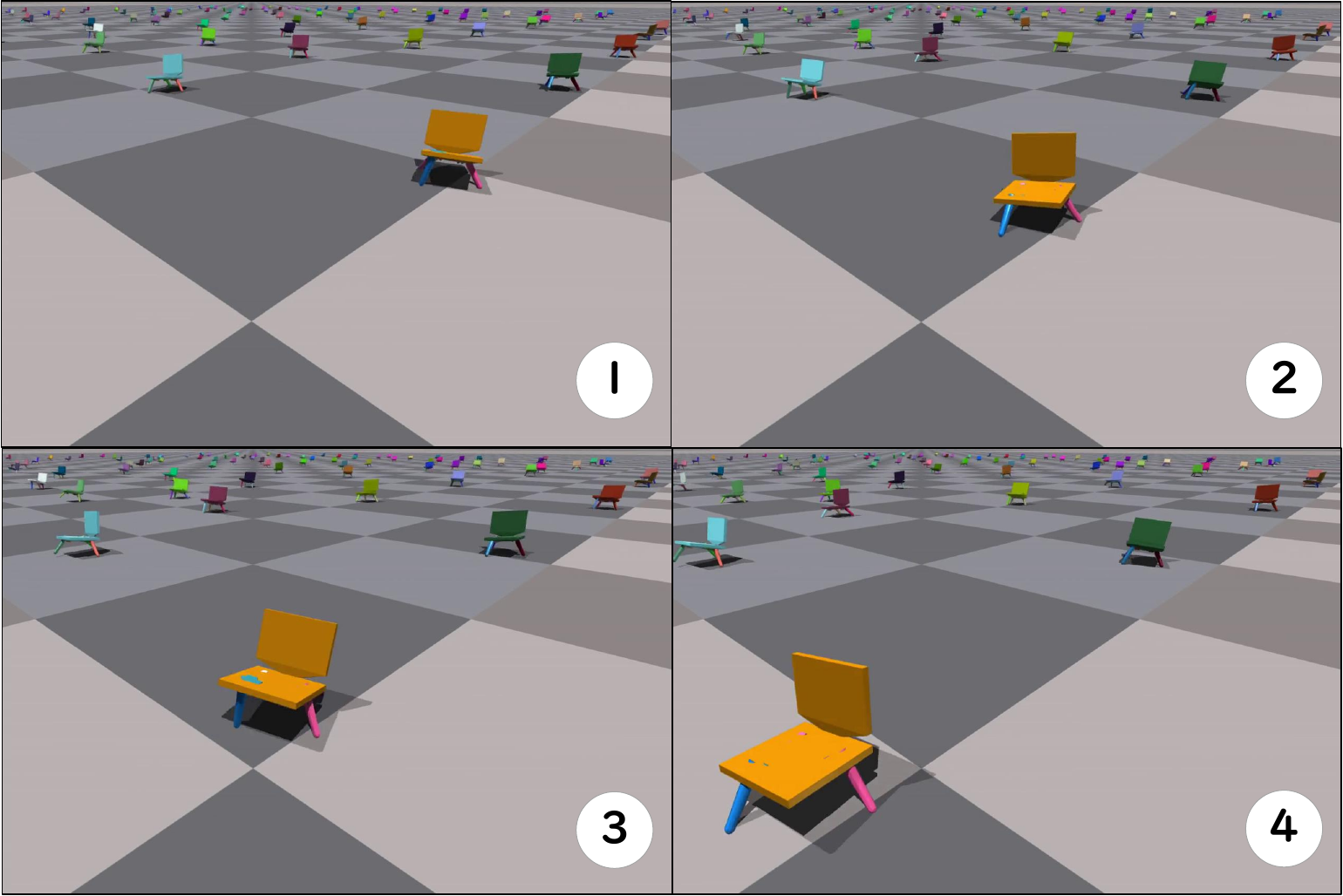}
    \caption{Walking motion on the simulator.}
  	\label{figure:issacgym_walk}
    \end{figure}
  
  \begin{table}[htb]
  	\centering
    \caption{Reset conditions for walking motion.}
  	\label{table:walk_reset_tab}
    \begin{tabular}{lp{6cm}lc} \hline
      Reset & Condition \\ \hline \hline
      max episode & When the episode exceeds 350. \\
      tilt & When $\|\bm{q}-[0,\ 0,\ 0,\ 1]^\top\| > 0.7$\\
      ground & When any corner of the seat surface contacts the ground. \\
      height & When the seat height becomes lower than \SI{5}{\milli\meter}. \\
      \end{tabular}
    \end{table}
  
  \renewcommand{\arraystretch}{1.5}
  \begin{table*}[htb]
  	\centering
    \caption{Rewards and weights in walking motion.}
  	\label{table:walk_reward_tab}
    \begin{tabular}{lp{14cm}lc} \hline
      Reward & Description and purpose & Weight \\ \hline \hline
      progress & $P - P_\text{pre}$ where $P=-\cfrac{\|\bm{p}_\text{target} - \bm{p}\|}{\text{dt}}$ and $P_\text{pre}$ is the previous cycle's $P$\par 
                 Reward for moving forward.& 30 \\
      height & $\min\left\{1, \cfrac{|\bm{p}|_z}{0.08}\right\}$\par Reward for high seat height. & 20 \\
      up & $\min\left\{1, \cfrac{u_\text{prj}}{0.93}\right\}$\par Reward for a vertically upward seat. & 5 \\
      heading & $\min\left\{1, \cfrac{1}{0.8}\bm{R_q}\bm{e}_x\cdot\frac{\bm{p}_\text{target} - \bm{p}}{\|\bm{p}_\text{target} - \bm{p}\|}\right\}$\par 
                Reward for the direction in which one is moving toward the target point. & 2 \\
      alive & Reward for not being reset. Unless it is reset, this gives a constant value. & 1 \\
      death & The reward is given when the reset condition other than the max episode is satisfied. This value overwrites the previous rewards. & -1 \\
      action & $\|\bm{a} - \bm{a}_\text{pre}\|^2$ where $\bm{a}_\text{pre}$ is the previous cycle's $\bm{a}$ \par Penalty for the absolute value of the difference of the command angle from the previous cycle. & -2 \\
      vel & $\left\|\cfrac{\bm{\omega}}{\omega_\text{max} - \omega_\text{tol}}\right\|^2$ where $\omega_{max} = 10.472,\ \omega_{tol} = 1$\par 
            Penalty for the magnitude of the joint angular velocity. & -2 \\
      \end{tabular}
    \end{table*}
  \renewcommand{\arraystretch}{1}
}%
{%
  歩行動作を強化学習させる際の報酬とその重みを\tabref{table:walk_reward_tab}に，リセット条件を\tabref{table:walk_reset_tab}にまとめる．
  座面を可能な限り水平で高く保ちながら前進したときに最も報酬が高くなるような設計になっている．
  一方でそのような状態ではないが報酬が高くなる場合について，背もたれを地面に擦りながら進んでいく歩容も見られたため，そのような歩容が学習されないようにリセット条件を設けている．
  131072体のエージェントを同時におよそ100epochの学習を行った後，
  センサ値や行動値などのパラメタにノイズをランダムに付与した上で追加でおよそ30epochの学習を行った
  学習後のシミュレータ上での歩行動作の様子を\figref{figure:issacgym_walk}に示す．

  \begin{figure}[htb]
  	\centering
  	\includegraphics[width=1\columnwidth]{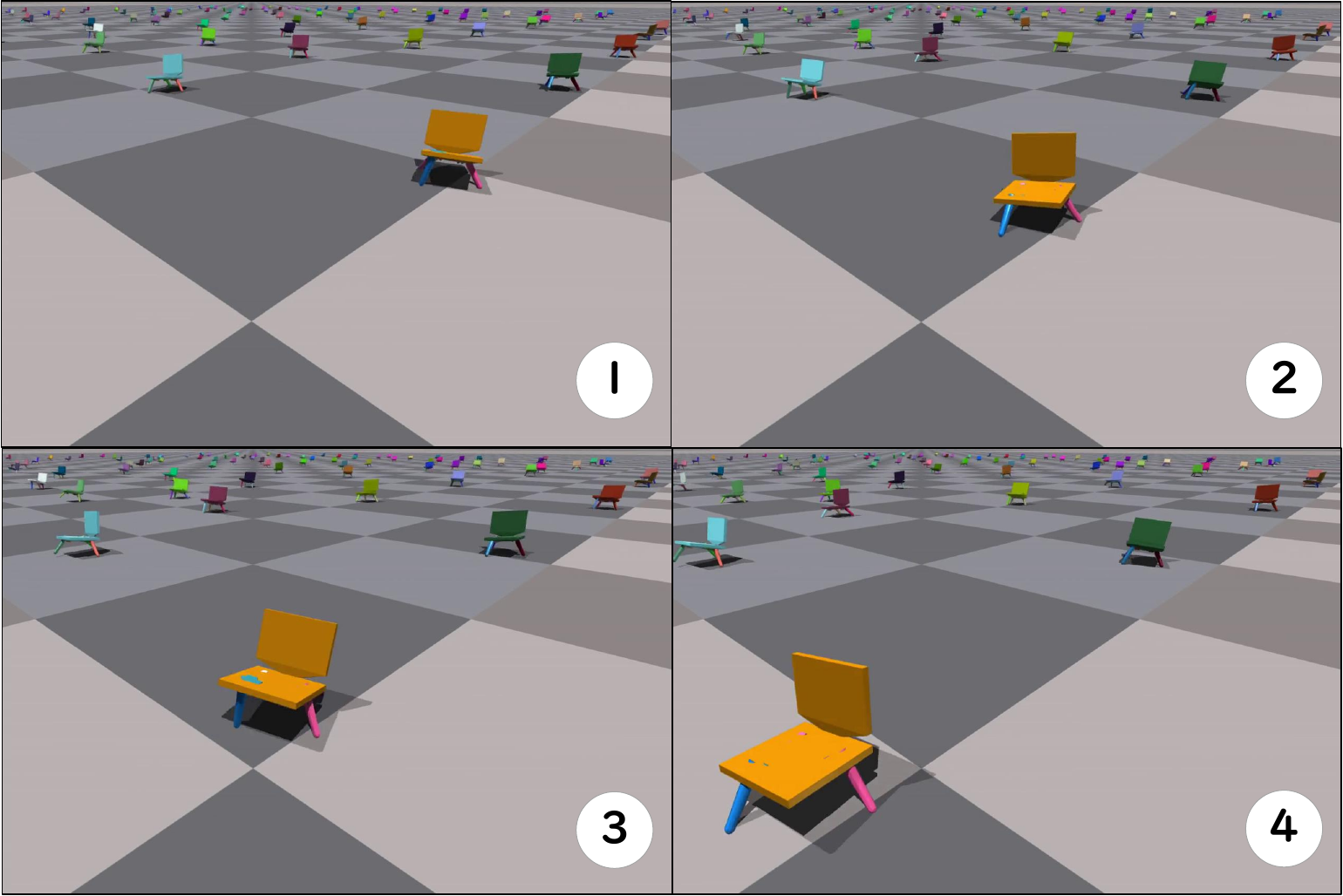}
    \caption{シミュレータ上での歩行動作.}
  	\label{figure:issacgym_walk}
    \end{figure}
  
  \begin{table}[htb]
  	\centering
    \caption{歩行動作におけるリセット条件.}
  	\label{table:walk_reset_tab}
    \begin{tabular}{lp{6cm}lc} \hline
      リセット & 条件 \\ \hline \hline
      max episode & episodeが350を超えたとき． \\
      tilt & $\|\bm{q}-[0,\ 0,\ 0,\ 1]^\top\| > 0.7$であるとき．\\
      ground & 座面のいずれかの角が地面に接触したとき． \\
      height & 座面高さが\SI{5}{\milli\meter}より低くなったとき． \\
      \end{tabular}
    \end{table}
  
  \renewcommand{\arraystretch}{1.5}
  \begin{table*}[t]
  	\centering
    \caption{歩行動作における報酬と重み.}
  	\label{table:walk_reward_tab}
    \begin{tabular}{lp{14cm}lc} \hline
      報酬 & 説明 & 重み \\ \hline \hline
      progress & $P - P_\text{pre}$ ただし，$P=-\cfrac{\|\bm{p}_\text{target} - \bm{p}\|}{\text{dt}}$であり$P_\text{pre}$は前周期の$P$\par 前進することに対する報酬． & 30 \\
      height & $\min\left\{1, \cfrac{|\bm{p}|_z}{0.08}\right\}$\par 座面高さが高いことに対する報酬．& 20 \\
      up & $\min\left\{1, \cfrac{u_\text{prj}}{0.93}\right\}$\par 座面が鉛直上向きになっていることに対する報酬． & 5 \\
      heading & $\min\left\{1, \cfrac{1}{0.8}\bm{R_q}\bm{e}_x\cdot\frac{\bm{p}_\text{target} - \bm{p}}{\|\bm{p}_\text{target} - \bm{p}\|}\right\}$\par 進んでいる方向が目標座標に向いていることに対する報酬．& 2 \\
      alive & リセットされていないことに対する報酬．リセットされない限りこれによって一定値が与えられる． & 1 \\
      death & max episode 以外のリセット条件を満たしたときに与えられる．これまでの報酬を上書きしてこの値となる． & -1 \\
      action & $\|\bm{a} - \bm{a}_\text{pre}\|^2$ ただし，$\bm{a}_\text{pre}$は前周期の$\bm{a}$\par 指令角度の前周期との差の絶対値に対する罰則． & -2 \\
      vel & $\left\|\cfrac{\bm{\omega}}{\omega_\text{max} - \omega_\text{tol}}\right\|^2$ ただし，$\omega_{max} = 10.472,\ \omega_{tol} = 1$\par 関節角速度の大きさに対する罰則． & -2 \\
      \end{tabular}
    \end{table*}
  \renewcommand{\arraystretch}{1}
}%

\subsubsection{Stand-up Motion} \label{subsubsec:rl_stand-up}
\switchlanguage%
{%
  The rewards and their weights are summarized in \tabref{table:stand_reward_tab} and the reset conditions are summarized in \tabref{table:stand_reset_tab}.
  The most reward is given to the agent that turns the seat face up, 
  but this is not enough to solve the problem of the agent that turns the seat face up with each leg folded inward and cannot stand up from there.
  The problem is solved by rewarding the outward spreading of each leg and resetting the agent that cannot stand up and whose seat surface has turned upward.
  131072 agents were simultaneously trained for approximately 250 epochs.
	Three initial postures were prepared for the robot at the beginning of training: right side, left side, and back on the ground, 
  and the robot learned to stand-up from these postures at once.
	The rise motion on the simulator after learning is shown in \figref{figure:issacgym_stand}.

	 \begin{figure}[htb]
  	\centering
  	\includegraphics[width=1\columnwidth]{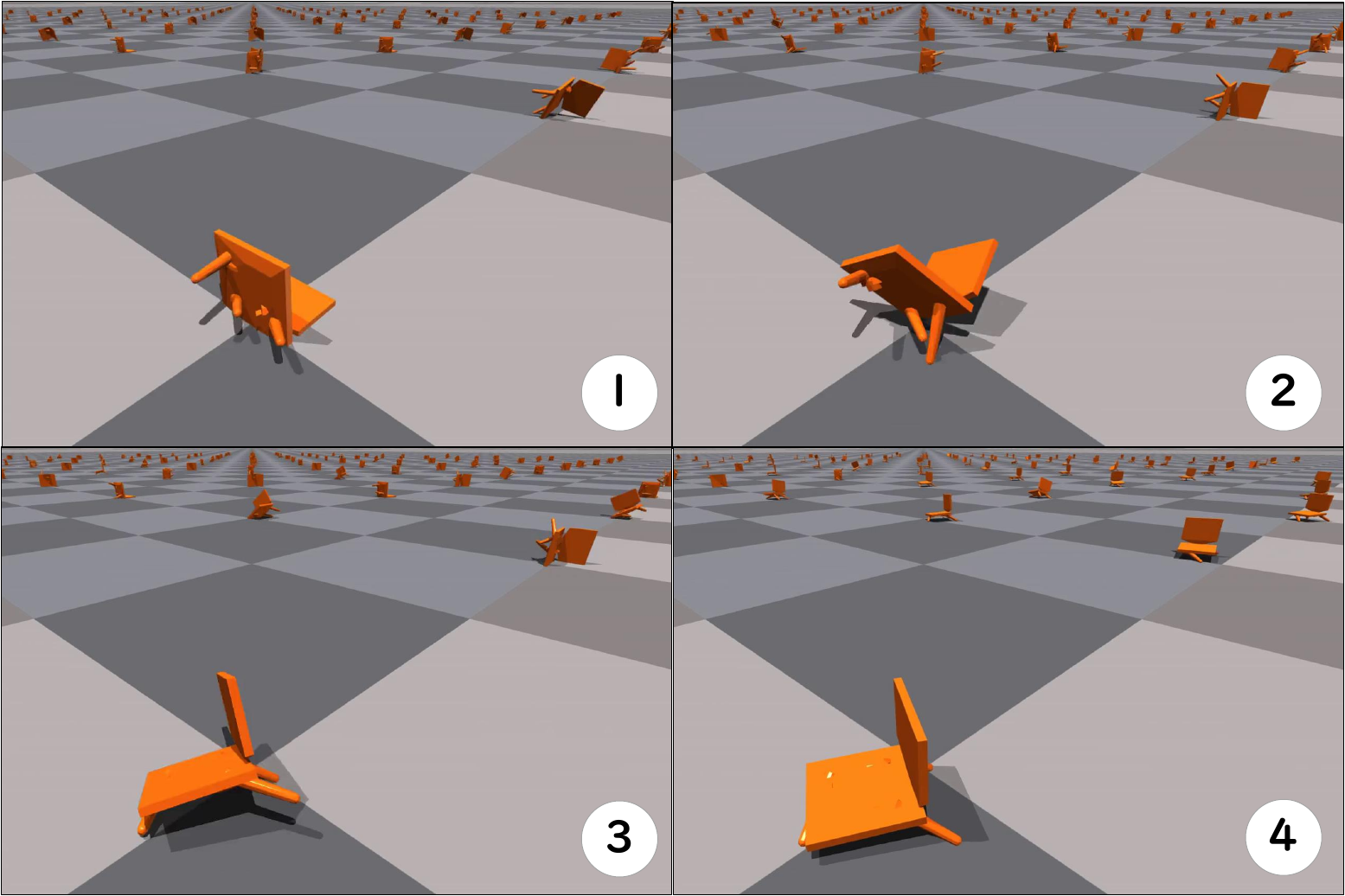}
    \caption{Stand-up motion on the simulator.}
  	\label{figure:issacgym_stand}
    \end{figure}
  
  \begin{table}[htb]
  	\centering
    \caption{Reset conditions in the stand-up motion.}
  	\label{table:stand_reset_tab}
    \begin{tabular}{lp{6cm}lc} \hline
      Reset & Condition \\ \hline \hline
      max episode & When the episode exceeds 350. \\
      flip & When $u_\text{prj} < -0.7$ \\
      fold & When $0.6 < u_\text{prj}$ and\par $\max\{\|[\theta_0,\ \theta_1,\ \theta_3,\ \theta_5 ]^\top-\bm{a}_\text{expand}\|_\infty\} > 1$ \\
      \end{tabular}
    \end{table}
  
  \renewcommand{\arraystretch}{1.5}
  \begin{table*}[htb]
  	\centering
    \caption{Rewards and weights in the stand-up motion.}
  	\label{table:stand_reward_tab}
    \begin{tabular}{lp{14cm}lc} \hline
      Reward & Description and purpose & Weight \\ \hline \hline
      up & $\min\{1, \exp\{2(u_\text{prj}-1)\}\}$\par Reward for the seat being vertically upward. & 250 \\
      standing & $\left\{\begin{array}{ll}
        \cfrac{1}{2|\arcsin\left(\min\{1, \|\bm{a}-\bm{a}_\text{stand}\|_4\}\right)| + 0.1} & (u_\text{prj} > 0.85) \\
      0 & (\text{otherwise}) \end{array}\right.$ where $\bm{a}_\text{stand} = [-0.1745,\ 0,\ -0.1745,\ 0,\ 0.1745,\ 0]^\top$
        \par Reward for the proximity of each joint angle to the standing posture. & 100 \\
      spreading & $\left\{\begin{array}{ll}
        \cfrac{1}{2|\arcsin\left(\min\{1, \|[\theta_0,\ \theta_1,\ \theta_3,\ \theta_5 ]^\top-\bm{a}_\text{expand}\|_4\}\right)| + 0.1} & (u_\text{prj} > 0.2) \\
      0 & (\text{otherwise}) \end{array}\right.$ where $\bm{a}_\text{expand} = [-1,\ -1,\ 1,\ -1]^\top$
        \par Reward for each leg not being folded inward during the process of seat upward. & 50 \\
      death & The reward is given when the reset condition other than the max episode is satisfied. This value overwrites the previous rewards. & -1 \\
      action & $\|\bm{a} - \bm{a}_\text{pre}\|^2$ where $\bm{a}_\text{pre}$ is the previous cycle's $\bm{a}$ \par Penalty for the absolute value of the difference of the command angle from the previous cycle. & -2 \\
      \end{tabular}
    \end{table*}
  \renewcommand{\arraystretch}{1}

}%
{%
  起き上がり動作を強化学習させる際の報酬とその重みを\tabref{table:stand_reward_tab}に，リセット条件を\tabref{table:stand_reset_tab}にまとめる．
  座面が上向きになることに対して最も報酬が与えられるようになっているが，それだけでは各脚が内側に折り畳まれたまま座面が上向きになり，そこからは立ち上がれない状態になってしまったため，
  各脚を外向き広げることに対して報酬を与えるとともに，立ち上がれない状態で座面が上向きなったエージェントをリセットすることでその問題を解決している．
  131072体のエージェントを同時におよそ250epochの学習を行った．
  学習始めのロボットの初期姿勢を3種類用意し，座面の右側面，左側面，背面が地面についた状態とすることで，それらの姿勢からの起き上がり動作を一度に学習した．
  学習後のシミュレータ上での起き上がり動作の様子を\figref{figure:issacgym_stand}に示す．

  \begin{figure}[htb]
  	\centering
  	\includegraphics[width=1\columnwidth]{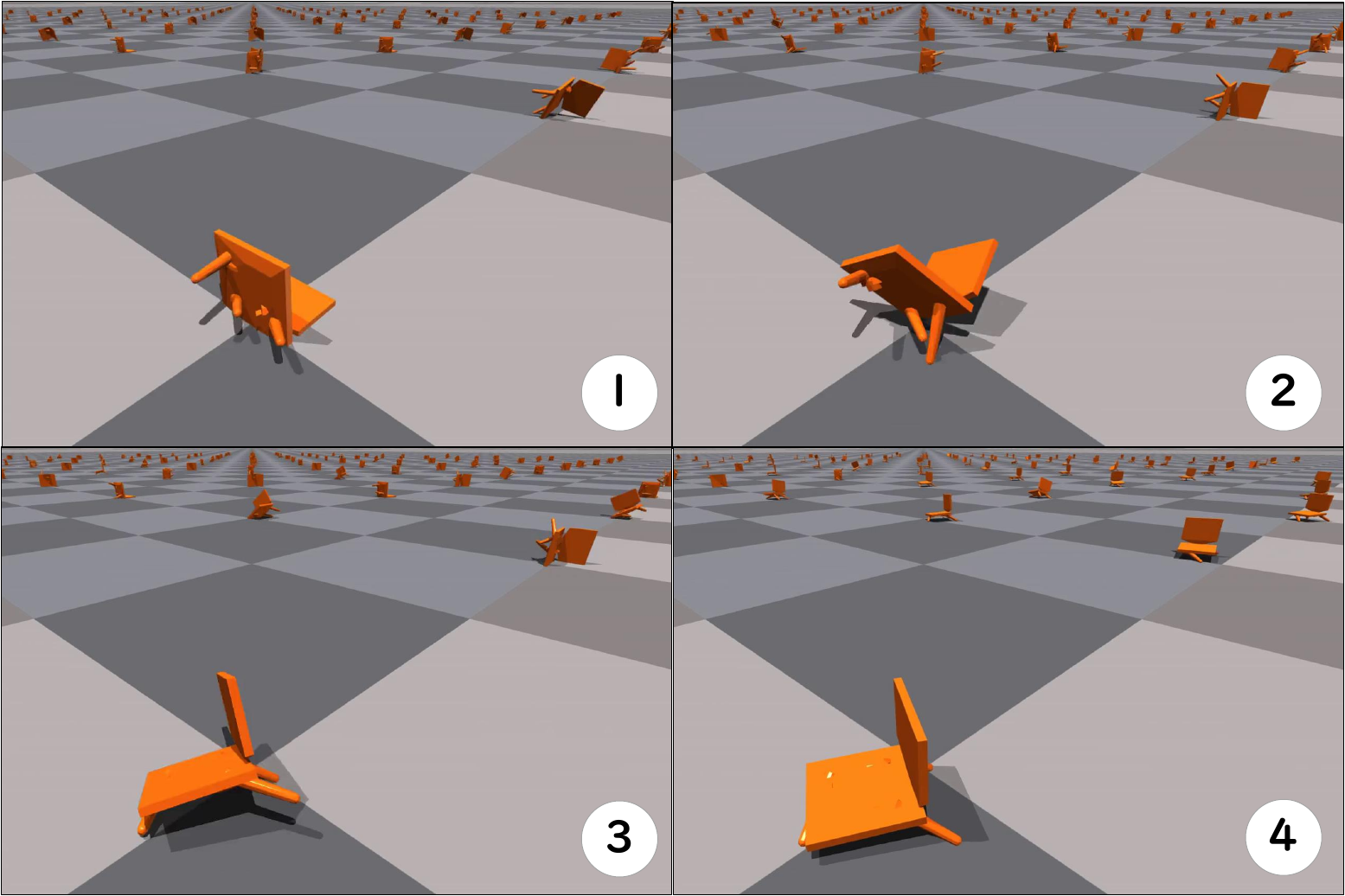}
    \caption{シミュレータ上での起き上がり動作.}
  	\label{figure:issacgym_stand}
    \end{figure}
  
  \begin{table}[htb]
  	\centering
    \caption{起き上がり動作におけるリセット条件.}
  	\label{table:stand_reset_tab}
    \begin{tabular}{lp{6cm}lc} \hline
      リセット & 条件 \\ \hline \hline
      max episode & episodeが350を超えたとき． \\
      flip &  $u_\text{prj} < -0.7$であるとき．\\
      fold & $0.6 < u_\text{prj}$ かつ\par $\max\{\|[\theta_0,\ \theta_1,\ \theta_3,\ \theta_5 ]^\top-\bm{a}_\text{expand}\|_\infty\} > 1$ \\
      \end{tabular}
    \end{table}
  
  \renewcommand{\arraystretch}{1.5}
  \begin{table*}[htb]
  	\centering
    \caption{起き上がり動作における報酬と重み.}
  	\label{table:stand_reward_tab}
    \begin{tabular}{lp{14cm}lc} \hline
      報酬 & 説明 & 重み \\ \hline \hline
      up & $\min\{1, \exp\{2(u_\text{prj}-1)\}\}$\par 座面が鉛直上向きになっていることに対する報酬．& 250 \\
      standing & $\left\{\begin{array}{ll}
        \cfrac{1}{2|\arcsin\left(\min\{1, \|\bm{a}-\bm{a}_\text{stand}\|_4\}\right)| + 0.1} & u_\text{prj} > 0.85 \\
      0 & \text{otherwise} \end{array}$ ただし，$\bm{a}_\text{stand} = [-0.1745,\ 0,\ -0.1745,\ 0,\ 0.1745,\ 0]^\top$
        \par 起立姿勢に各関節角度が近いことに対する報酬．& 100 \\
      expanding & $\left\{\begin{array}{ll}
        \cfrac{1}{2|\arcsin\left(\min\{1, \|[\theta_0,\ \theta_1,\ \theta_3,\ \theta_5 ]^\top-\bm{a}_\text{expand}\|_4\}\right)| + 0.1} & u_\text{prj} > 0.2 \\
      0 & \text{otherwise} \end{array}$ ただし，$\bm{a}_\text{expand} = [-1,\ -1,\ 1,\ -1]^\top$
        \par 座面が上向きになる過程で，各脚が内側に折り畳まれていないことに対する報酬． & 50 \\
      death & max episode 以外のリセット条件を満たしたときに与えられる．これまでの報酬を上書きしてこの値となる． & -1 \\
      action & $\|\bm{a} - \bm{a}_\text{pre}\|^2$ ただし，$\bm{a}_\text{pre}$は前周期の$\bm{a}$\par 指令角度の前周期との差の絶対値に対する罰則． & -2 \\
      \end{tabular}
    \end{table*}
  \renewcommand{\arraystretch}{1}
}%

\section{Experiments} \label{sec:experiments}
\subsection{Experiment Overview and Environment} \label{subsec:experimental-condition}
\switchlanguage%
{%
	
  \begin{figure}[htb]
  	\centering
  	\includegraphics[width=1\columnwidth]{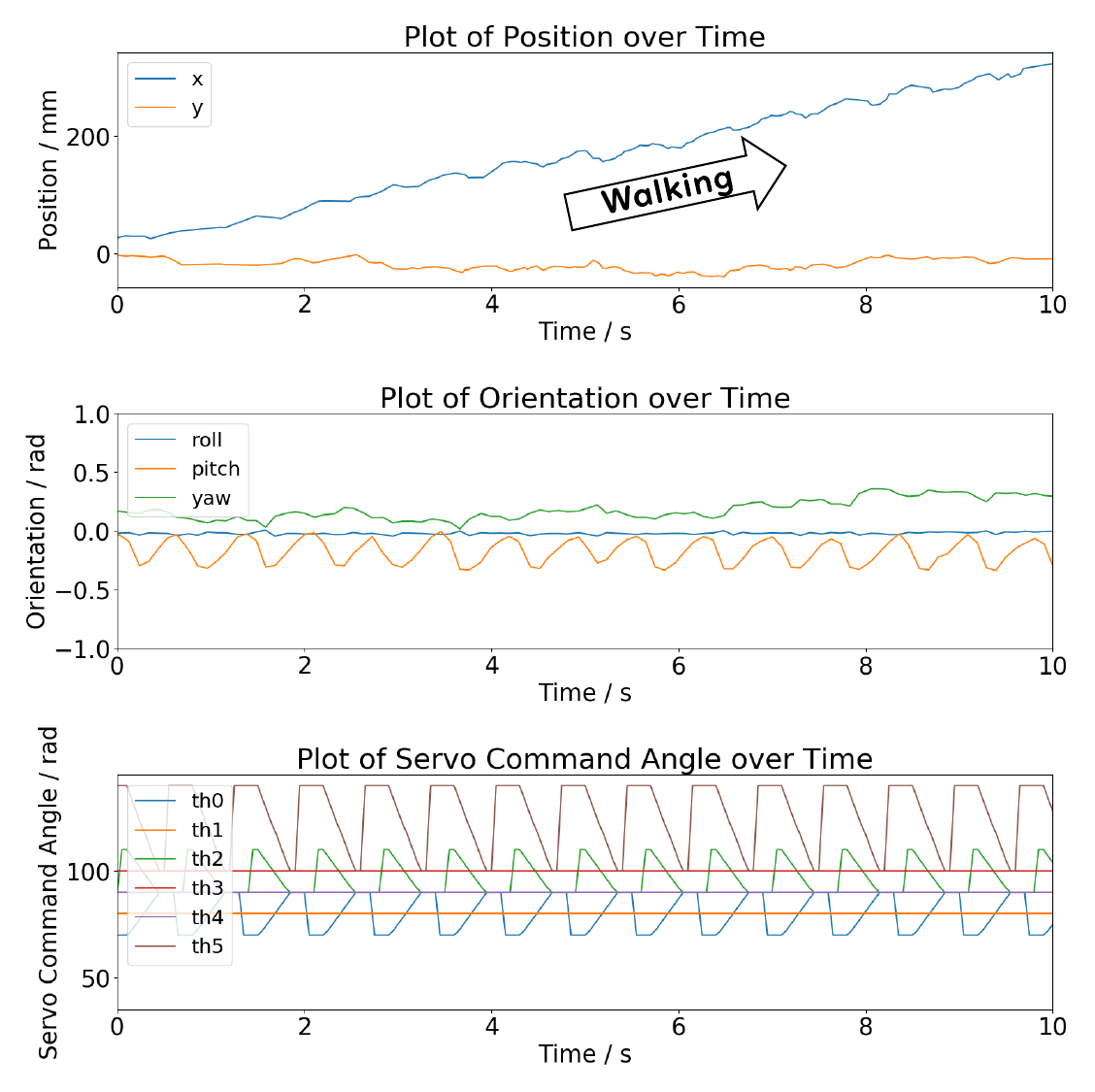}
    \caption{Result of walking motion by connecting essential postures.}
  	\label{figure:handmade_walk_fig}
    \end{figure}

  \begin{figure}[htb]
  	\centering
  	\includegraphics[width=1\columnwidth]{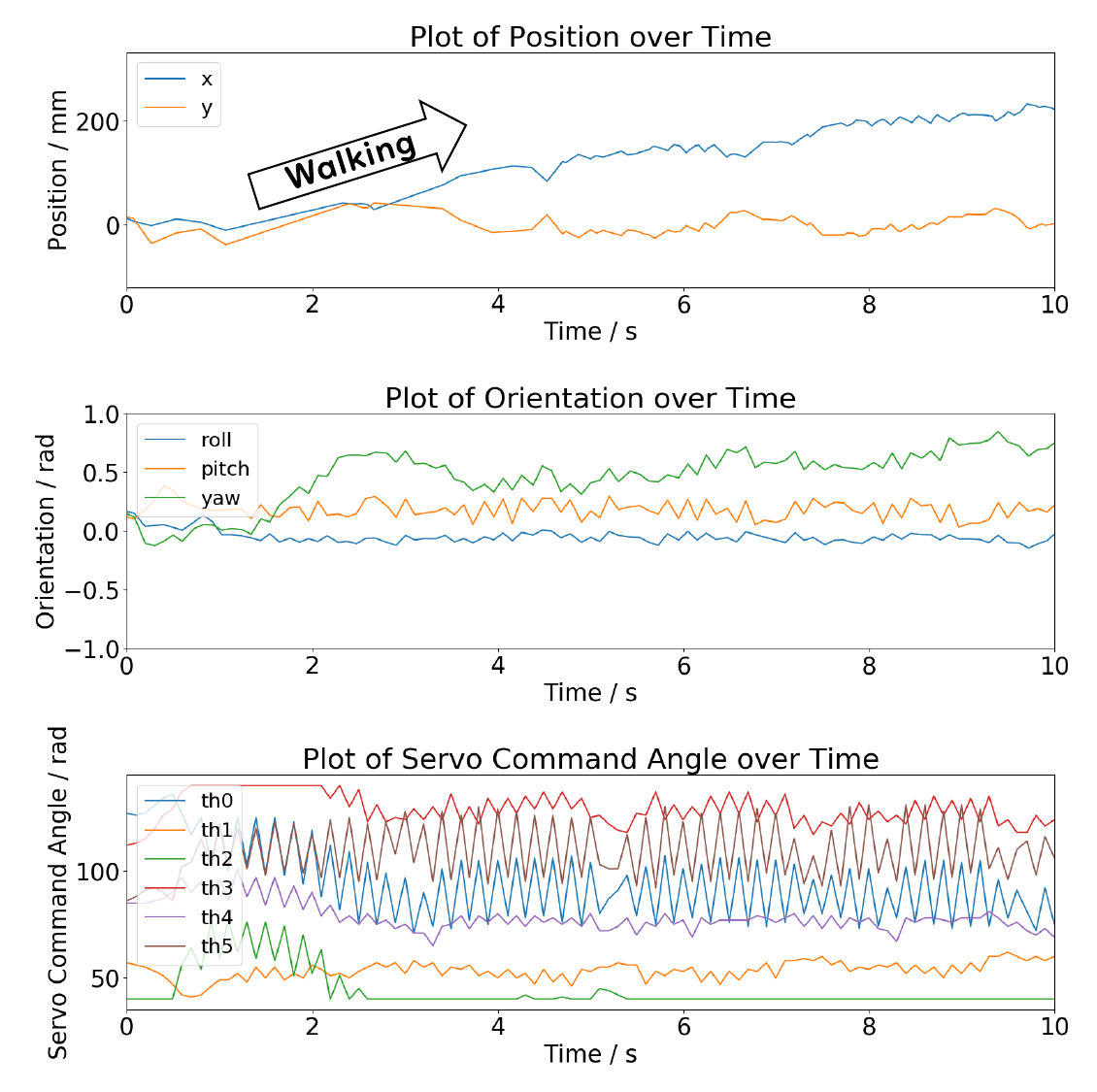}
    \caption{Results of walking motion by reinforcement learning.}
  	\label{figure:rl_walk_fig}
    \end{figure}

  In this experiment, the $x, y$ coordinates, roll angle, pitch angle, and yaw angle of the robot are measured, 
  and the command angles of the servo motors in the walking and stand-up motions are measured over time 
  to analyze the gait by connecting essential postures and the gait by the reinforcement learning method.
  During walking, the robot moves forward along the $x$-axis, and during stand-up motion, the robot stands up from a state in which the side of the robot's seat is on the ground.
  The experimental environment is shown in \figref{figure:experiment_env}.
  For evaluation purposes, the robot's $x, y$ coordinates are measured using the AR marker, and the roll, pitch, yaw angles are measured using the IMU .
}%
{%
  \begin{figure}[htb]
  	\centering
  	\includegraphics[width=0.7\columnwidth]{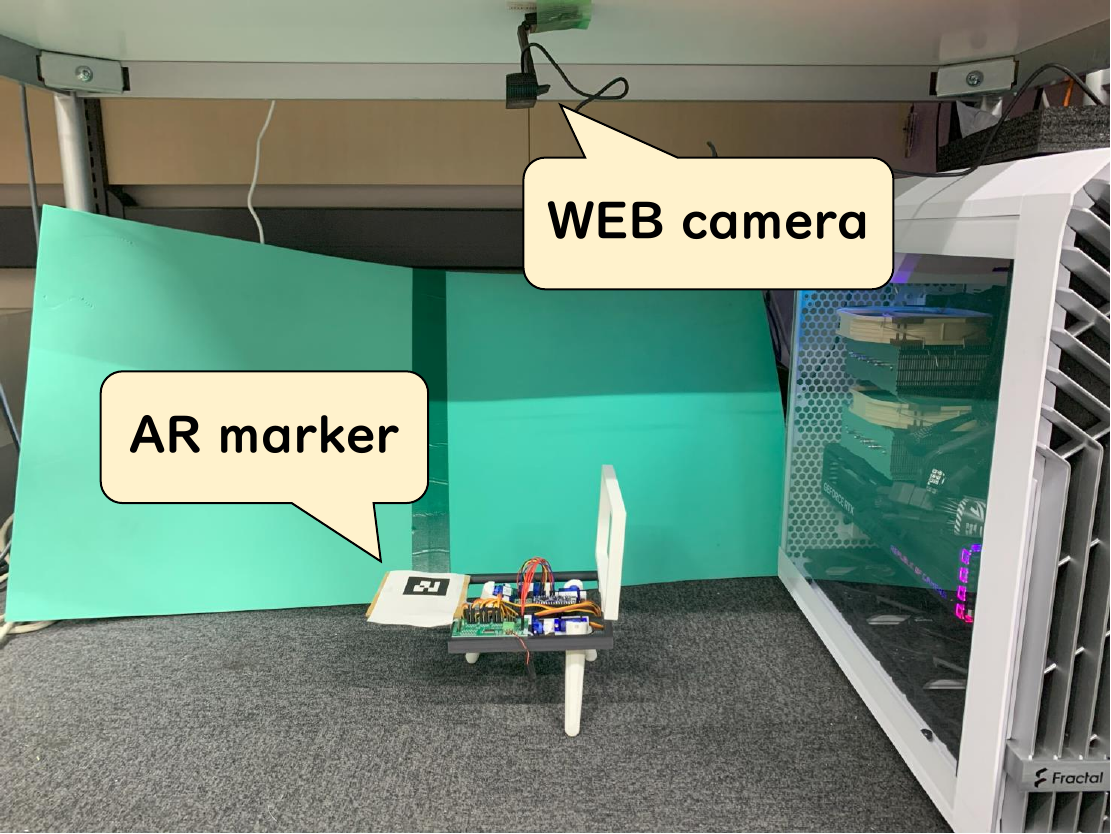}
  	\caption{実験環境.}
  	\label{figure:experiment_env}
    \end{figure}
  
  歩行動作，起き上がり動作でのロボットのx，y座標およびロール，ピッチ，ヨー角，サーボモータ指令角度を時系列で計測する実験を行うことにより，
	姿勢遷移による発見的な歩容および強化学習による歩容をそれぞれ計測し比較する．
  歩行動作ではロボットをx軸に沿って前に進ませる動作，起き上がり動作では，ロボットの座面側面が地面についている状態から起き上がるという動作を行う．
  実験環境は\figref{figure:experiment_env}に示す通りで，ロボットにARマーカーを貼り付け，それを上からWEBカメラで写すことで，
  ロボットのx，y座標をARマーカーで計測し，ロール，ピッチ，ヨー角はIMUから計測する．
}%

\subsection{Walking Motion}
\switchlanguage%
{%
  \figref{figure:handmade_walk_fig} is the result of walking motion by connecting essential postures.
  The walking motion is realized by repeatedly transitioning from the posture $\bm{S}_0$ to $\bm{S}_f$.
  Since the posture transition is a feed-forward command, the command angle shows always the same pattern regardless of the robot's position or posture.
  The $x, y$ coordinates of the graph show that the robot is moving forward at a constant speed.
  The pitch angle shows that the robot is swinging its posture in the pitch direction at a constant period.

  \figref{figure:rl_walk_fig} is the result of the walking motion by reinforcement learning.
  The $x, y$ coordinates of the graph show that the robot is moving forward, but its speed is not constant.
  The robot moves forward significantly from time 2 to 4 seconds, while it hardly moves forward at time 8 to 10 seconds.
  The yaw angle shows that the robot is moving forward while looking at $\ang{35}$ diagonally to the left.
  The gait by reinforcement learning was a gait in which the entire body was shaken up and down by vibrating the joint angles. 
  When the body was about to move in the upward direction by the vibration, the gait was to move forward by kicking the ground with the legs.
  The pitch angle was always upward and the amplitude of the vibration was smaller than that of walking motion by connecting essential postures.
}%
{%
  \figref{figure:handmade_walk_fig}は姿勢遷移による発見的な歩行動作の結果である．
  姿勢$\bm{S}_0$から$\bm{S}_f$へ遷移させることを繰り返すことにより歩行動作を実現している．
  姿勢遷移はフィードフォワード的な指令であるため，指令角度はロボットの位置や姿勢によらず常に同じものが繰り返し指令されている．
  グラフのx，y座標を見ると，ロボットが前向きにおよそ一定の速度で進んでいることが見られる．
  また，ピッチ角に注目すると，ロボットはおよそ一定の周期でピッチ方向に姿勢を揺らしながら歩行していることがわかる．
  
  \begin{figure}[htb]
  	\centering
  	\includegraphics[width=1\columnwidth]{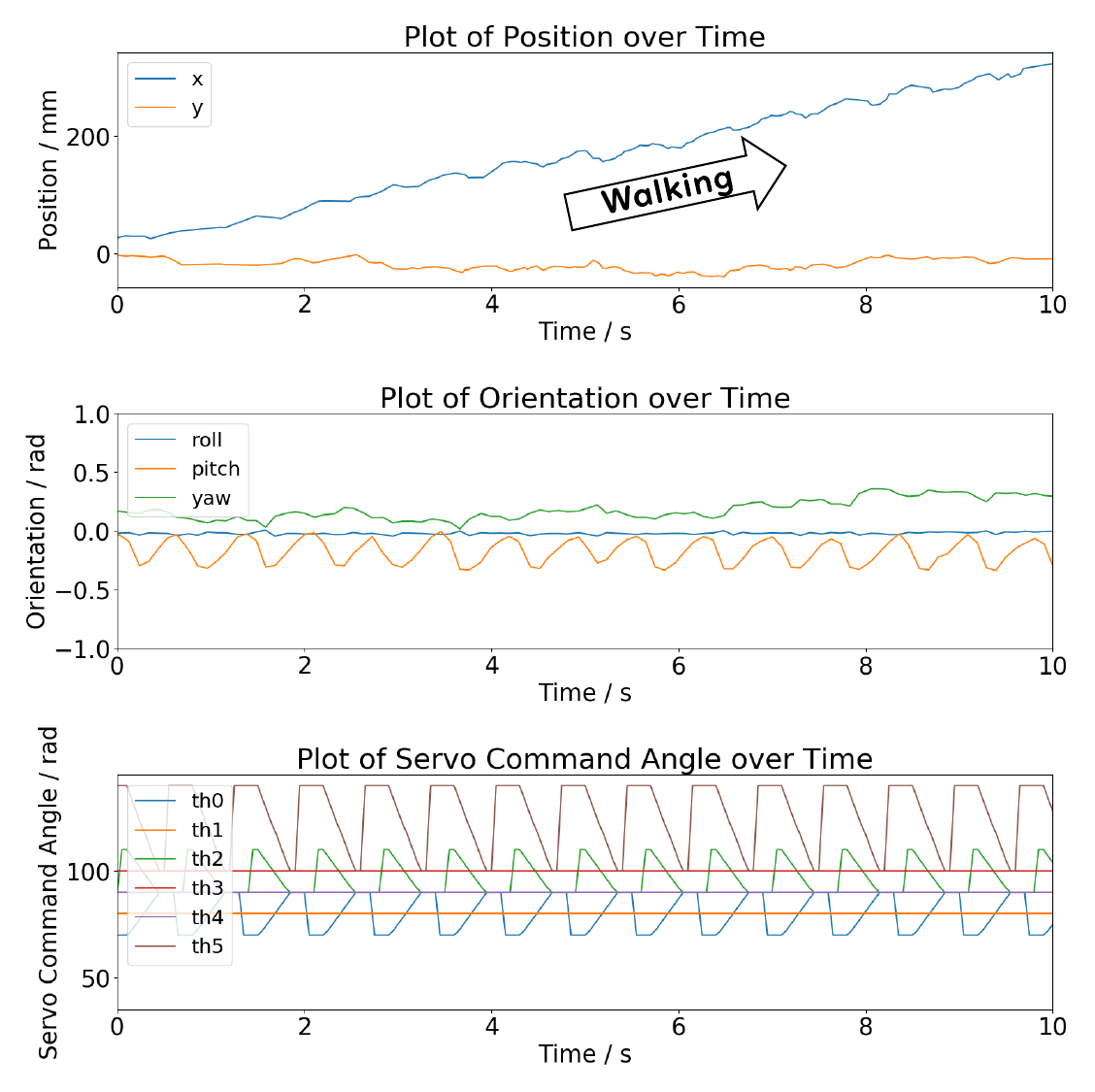}
    \caption{姿勢遷移による発見的な歩行動作の結果.}
  	\label{figure:handmade_walk_fig}
    \end{figure}
  
  \figref{figure:rl_walk_fig}は強化学習による歩行動作の結果である．
  グラフのx，y座標をみると，ロボットは前向きに進んでいることがわかるが，その速度は一定ではなくまばらであり，
	時刻2秒から4秒あたりに大きく前に進んでいる一方で時刻8秒から10秒ではほとんどロボットは前に進めていない．
  またヨー角を見ると，ロボットは左斜め前である$\ang{35}$あたりを向きながら前に進んでいるということがわかる．
  強化学習による歩容は，関節角度を振動させることによって体全体を上下に揺らし，
	その振動で上方向に体が向かおうとするタイミングにおいて，脚で地面を蹴ることで前に進もうという歩容であった．
  また，ピッチ角は発見的な歩行動作に比べると常に上向きで振動の振幅が小さいことが見て取れる．
  
  \begin{figure}[htb]
  	\centering
  	\includegraphics[width=1\columnwidth]{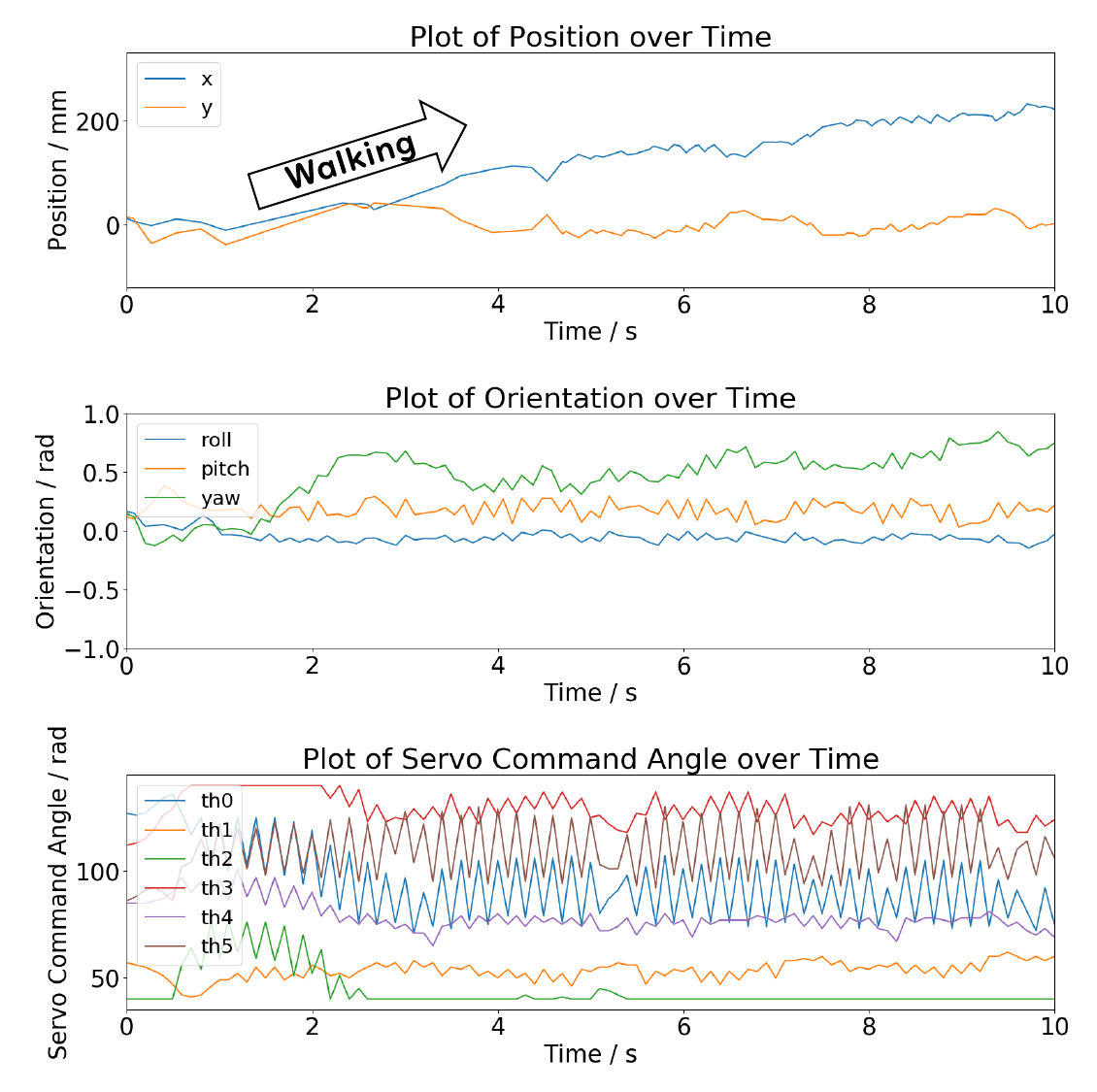}
    \caption{強化学習による歩行動作の結果.}
  	\label{figure:rl_walk_fig}
    \end{figure}
}%

\subsection{Stand-up Motion}
\switchlanguage%
{%
  \begin{figure}[htb]
  	\centering
  	\includegraphics[width=1\columnwidth]{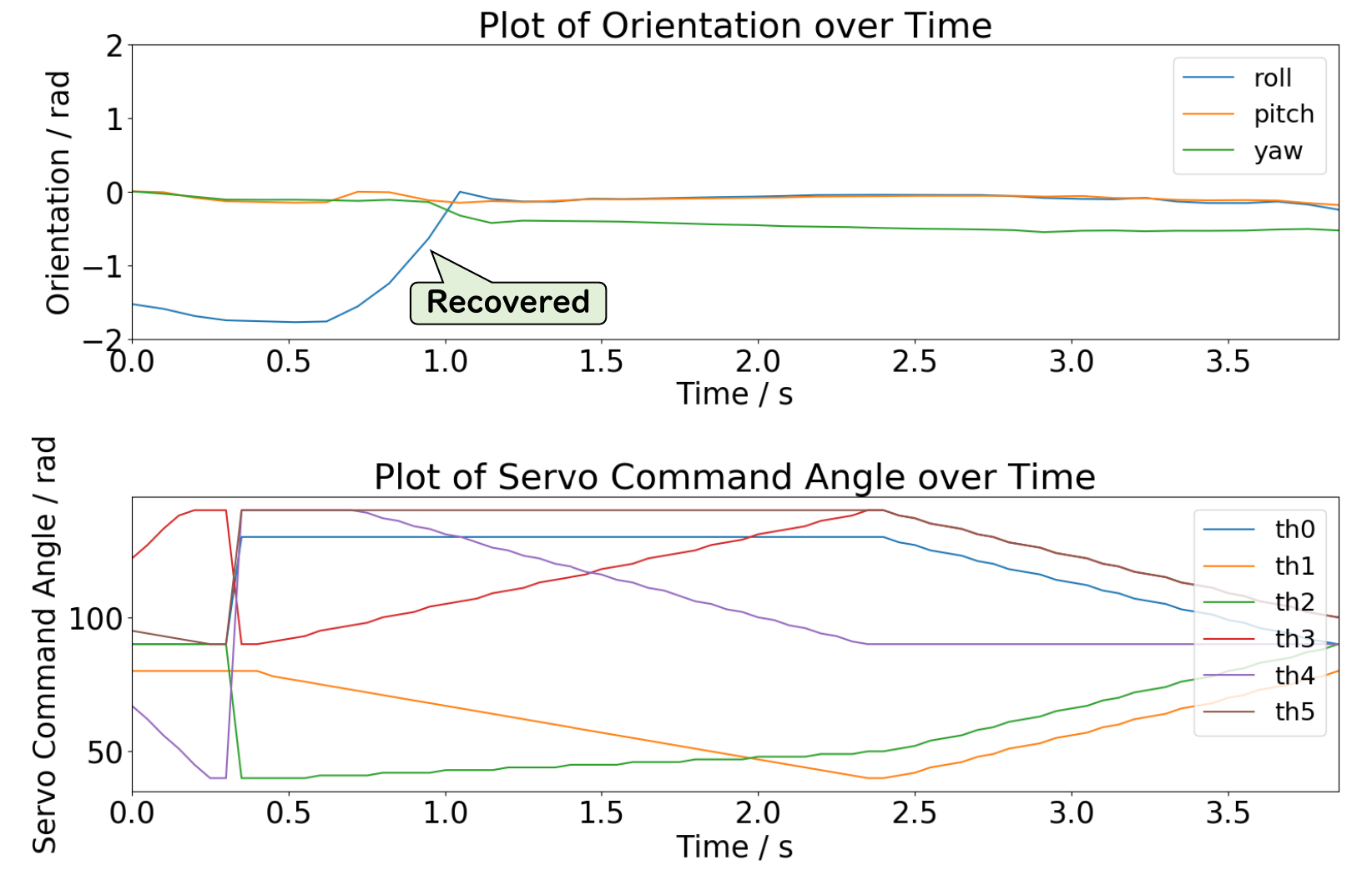}
    \caption{The result of stand-up by connecting essential postures.}
  	\label{figure:handmade_stand_fig}
    \end{figure}

  \begin{figure}[htb]
  	\centering
  	\includegraphics[width=1\columnwidth]{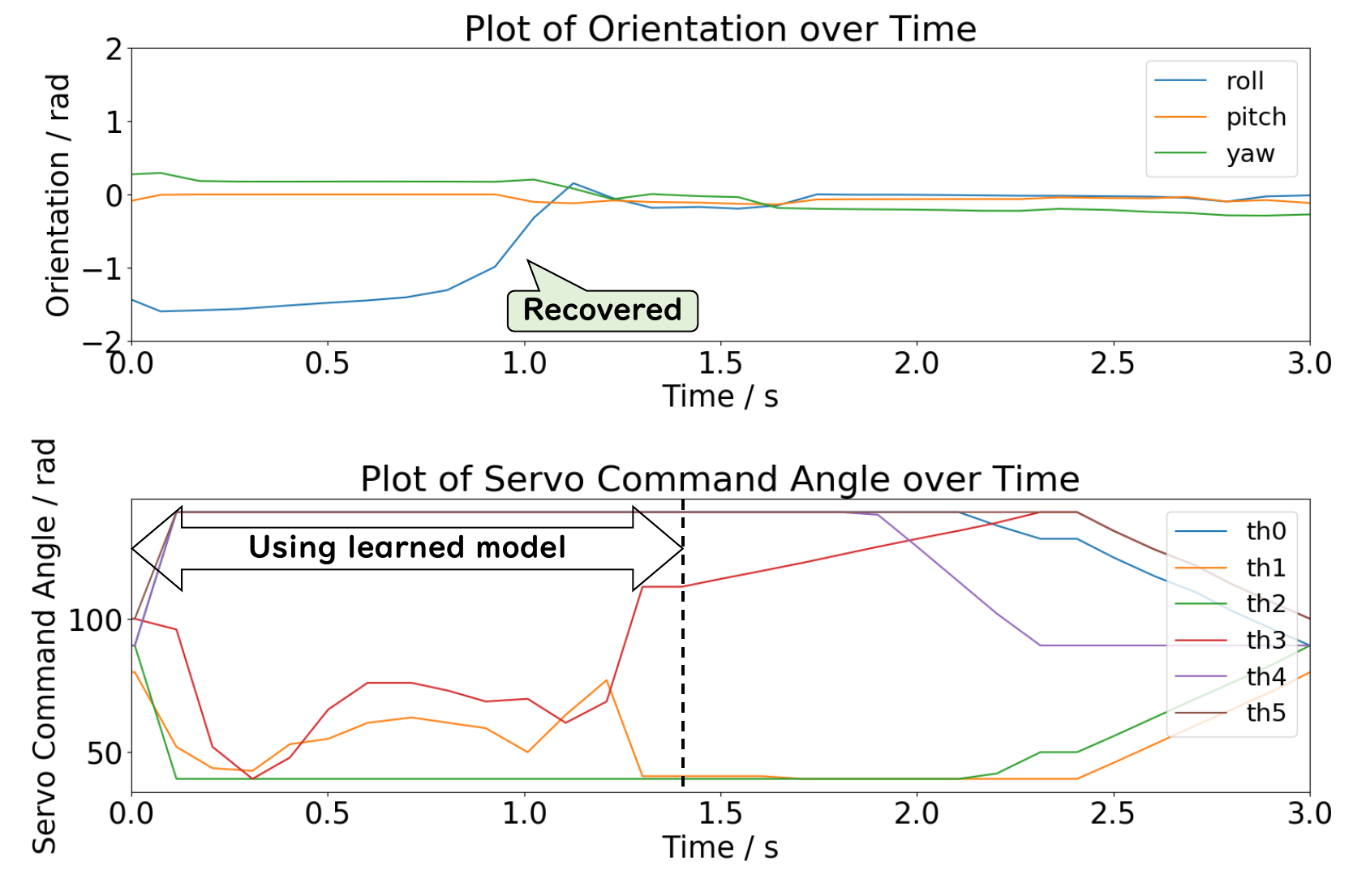}
    \caption{Result of stand-up by reinforcement learning.}
  	\label{figure:rl_stand_fig}
    \end{figure}

  \figref{figure:handmade_stand_fig} is the result of a stand-up motion by connecting essential postures.
  The robot is performing the action of stand-up from the position where the right side of the seat is on the ground.
	The roll angle shows that the robot is tilted further than $\ang{90}$ and then rises to $\ang{0}$.
	The servo motor command angle shows that the robot is pushing and tilting its body with its legs and then quickly releasing it to raise its posture.

  \figref{figure:rl_stand_fig} shows the result of stand-up motion by reinforcement learning.
  The movement from the right side of the seat to stand-up is the result of reinforcement learning, and the movements after that are pre-defined movements.
	As with the stand-up motion by connecting essential postures, the seat was tilted backward and then started to stand-up.
	In additon, the $\theta_1$ and $\theta_3$ change significantly from 1 second to 1.5 seconds. 
  $\theta_3$ is the leg that kicks the ground, while $\theta_1$ is the upper leg of the pair. 
	The large change in $\theta_1$ generates inertial force in the direction of stand-up, which is also utilized to realize the stand-up motion.

  As described in the \secref{subsubsec:rl_stand-up}, 
  the initial posture of the robot at the beginning of the learning process is 
  not only the robot's right side of the seat on the ground, but also the left side and the back of the robot.
	Therefore, the robot can rise from various postures.
	This is shown in \figref{figure:rl_stand_many_fig}, which shows the robot in a standing posture that is repeatedly knocked over by a hand from the outside, 
  and then stand-up from that posture.
	The robot was knocked over in the order of right side, back, left side, and right side, and it can be seen that the robot rises from any of the postures.
	Although the yaw angle was included in the observation, it was not involved in the rewards, so the robot took various postures with no restriction on the yaw angle.

  \begin{figure}[htb]
  	\centering
  	\includegraphics[width=1\columnwidth]{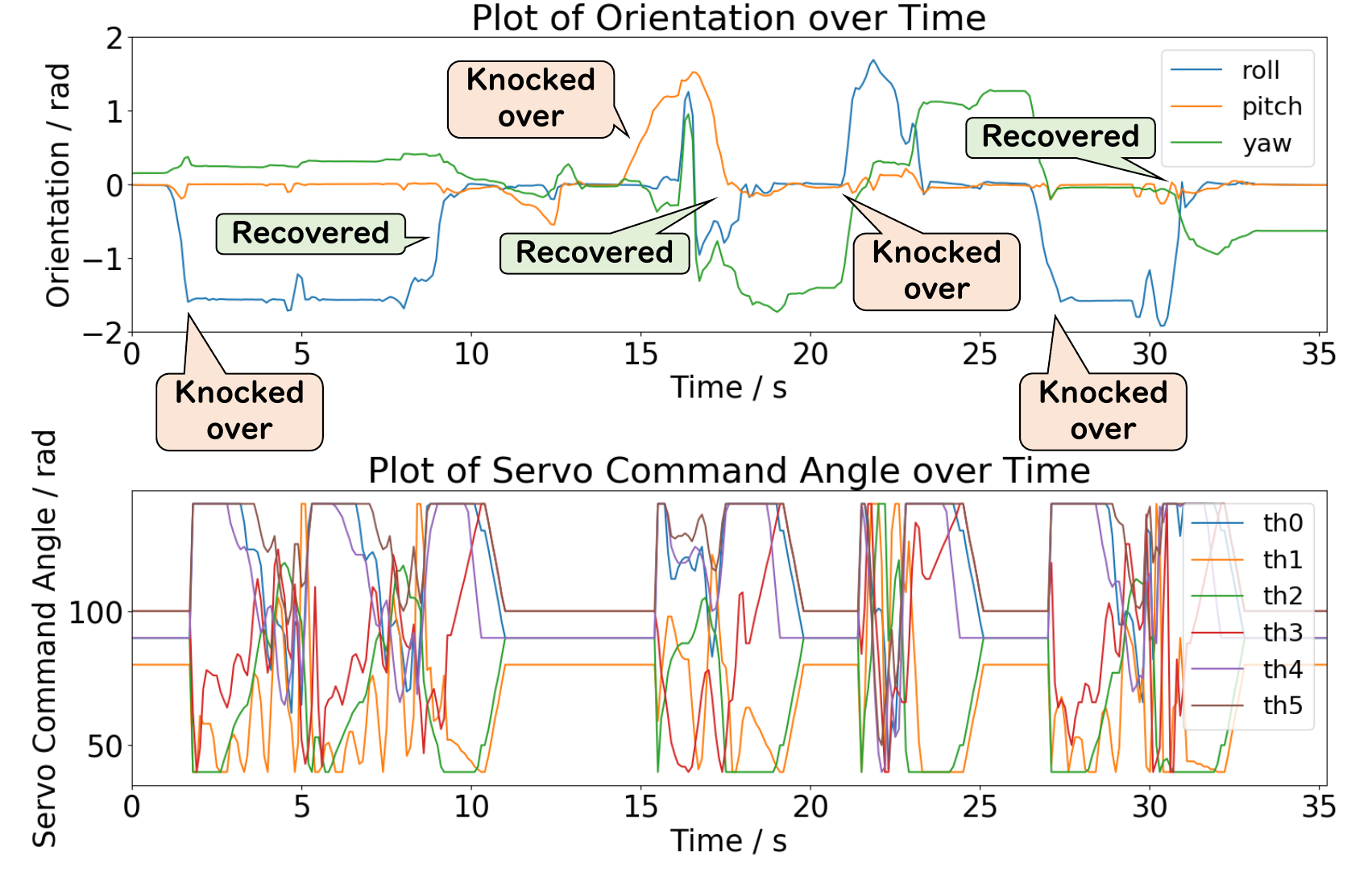}
    \caption{Stand-up from various postures by reinforcement learning.}
  	\label{figure:rl_stand_many_fig}
    \end{figure}

  \begin{figure}[htb]
  	\centering
  	\includegraphics[width=0.7\columnwidth]{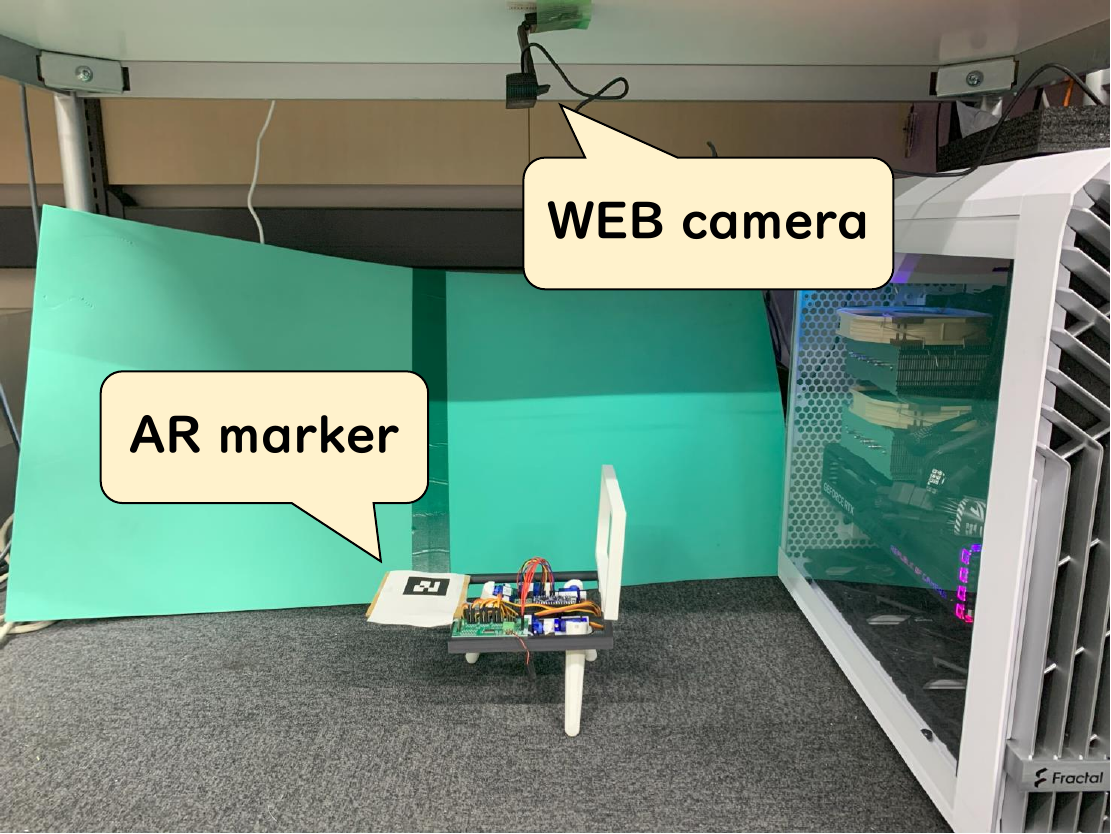}
  	\caption{Experimental environment.}
  	\label{figure:experiment_env}
    \end{figure}

}%
{%
  \figref{figure:handmade_stand_fig}は姿勢遷移による発見的な起き上がり動作の結果である．
  座面の右側面が地面についた状態から起き上がるという動作を実行している．
  ロール角に注目すると，一度$\ang{90}$よりさらに傾いた後，$\ang{0}$に起き上がっていることがわかる．
  サーボモータ指令角度で見ると，脚で体を押し傾けてから素早くそれを解放することで姿勢を起き上がらせていることがわかる．
  
  \begin{figure}[htb]
  	\centering
  	\includegraphics[width=1\columnwidth]{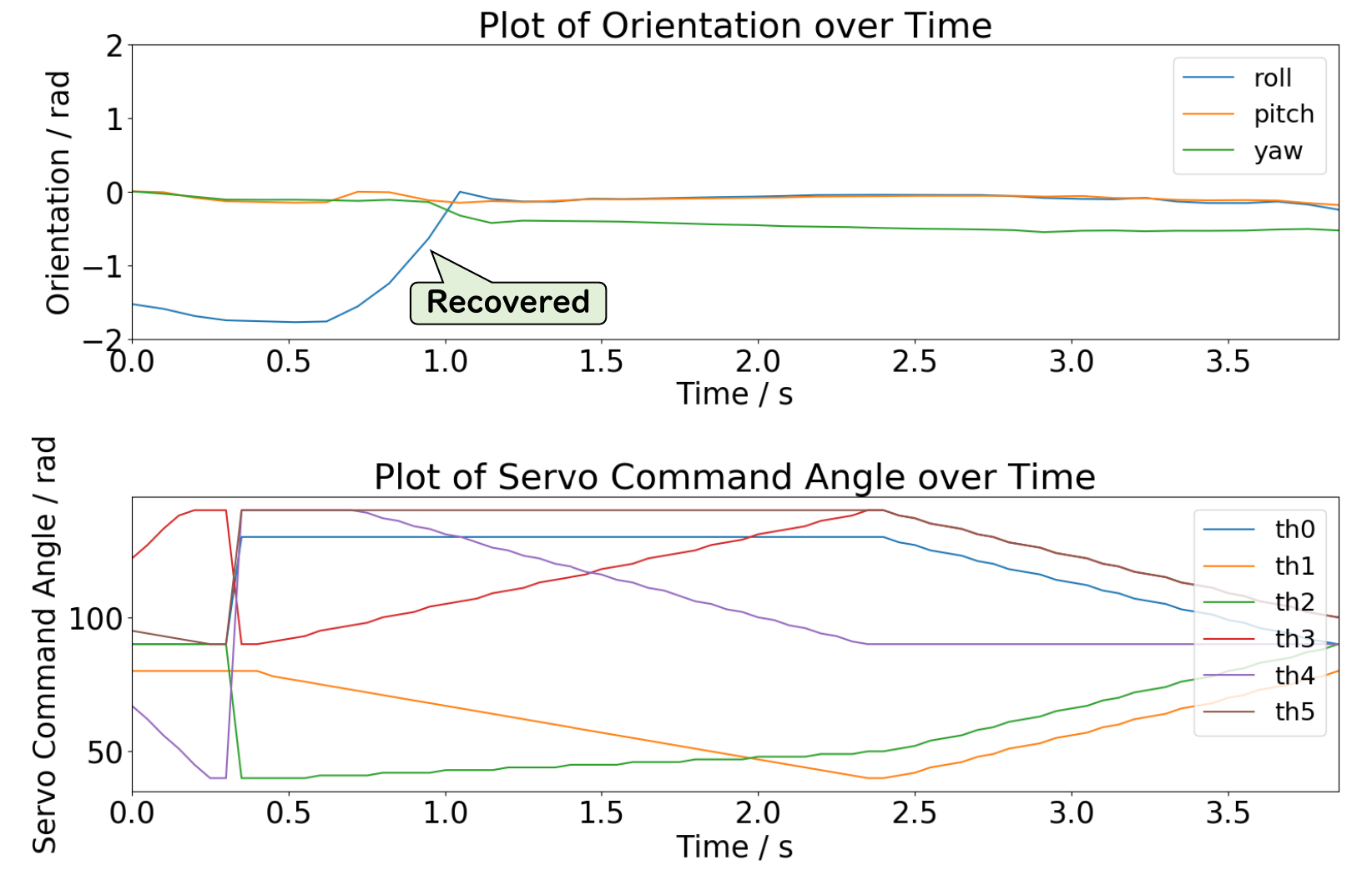}
    \caption{姿勢遷移による発見的な起き上がりの結果.}
  	\label{figure:handmade_stand_fig}
    \end{figure}
  
  \figref{figure:rl_stand_fig}は強化学習による起き上がり動作の結果である．
  座面の右側面がついた状態から起き上がるまでが強化学習によるものであり， それ以降は予め指定した動作である．
  強化学習による起き上がり動作も発見的なものと同様，一度，座面を逆に傾けた上で起き上がりを始めている．
  また，時刻1秒から1.5秒において$\theta_1$と$\theta_3$が大きく変化しているが，$\theta_3$は地面を蹴る脚である一方で，
  $\theta_1$はそれと対の上側の脚であり，$\theta_1$を大きく変化させていることは起き上がる方向へ慣性力を発生させることになり，
  それも利用して起き上がり動作を実現してることがわかる．
  
  \begin{figure}[htb]
  	\centering
  	\includegraphics[width=1\columnwidth]{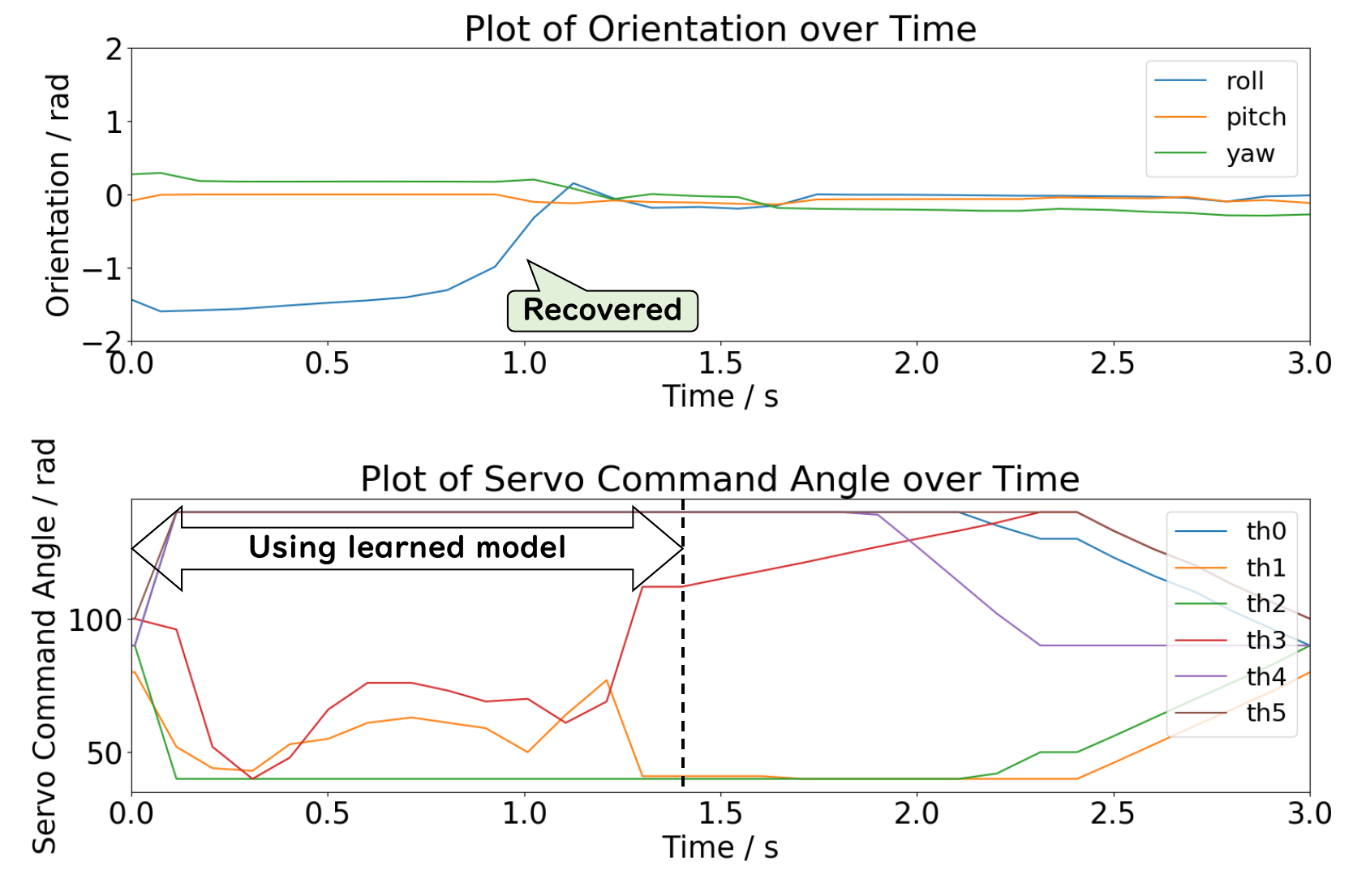}
    \caption{強化学習による起き上がりの結果.}
  	\label{figure:rl_stand_fig}
    \end{figure}
  
  また強化学習による起き上がりでは，その歩容生成手法でも述べたとおり，学習始めのロボットの初期姿勢を座面の右側面が地面についた状態のみならず，
  左側面，背面も含まれるので，様々な姿勢から起き上がり動作をすることが可能である．
  それを示すものが\figref{figure:rl_stand_many_fig}であり，起立姿勢にあるロボットを外部から手によって倒し，その状態から起き上がるというものを繰り返したものである．
  右側面，背面，左側面，右側面の順で倒れさせているが，いずれの姿勢からもロボットが起き上がっていることがわかる．
  学習の際，観測にヨー角は入ってはいるものの，報酬に関与していないため，ヨー角に関しては特に限定されない様々な姿勢をとっている．
  
  \begin{figure}[htb]
  	\centering
  	\includegraphics[width=1\columnwidth]{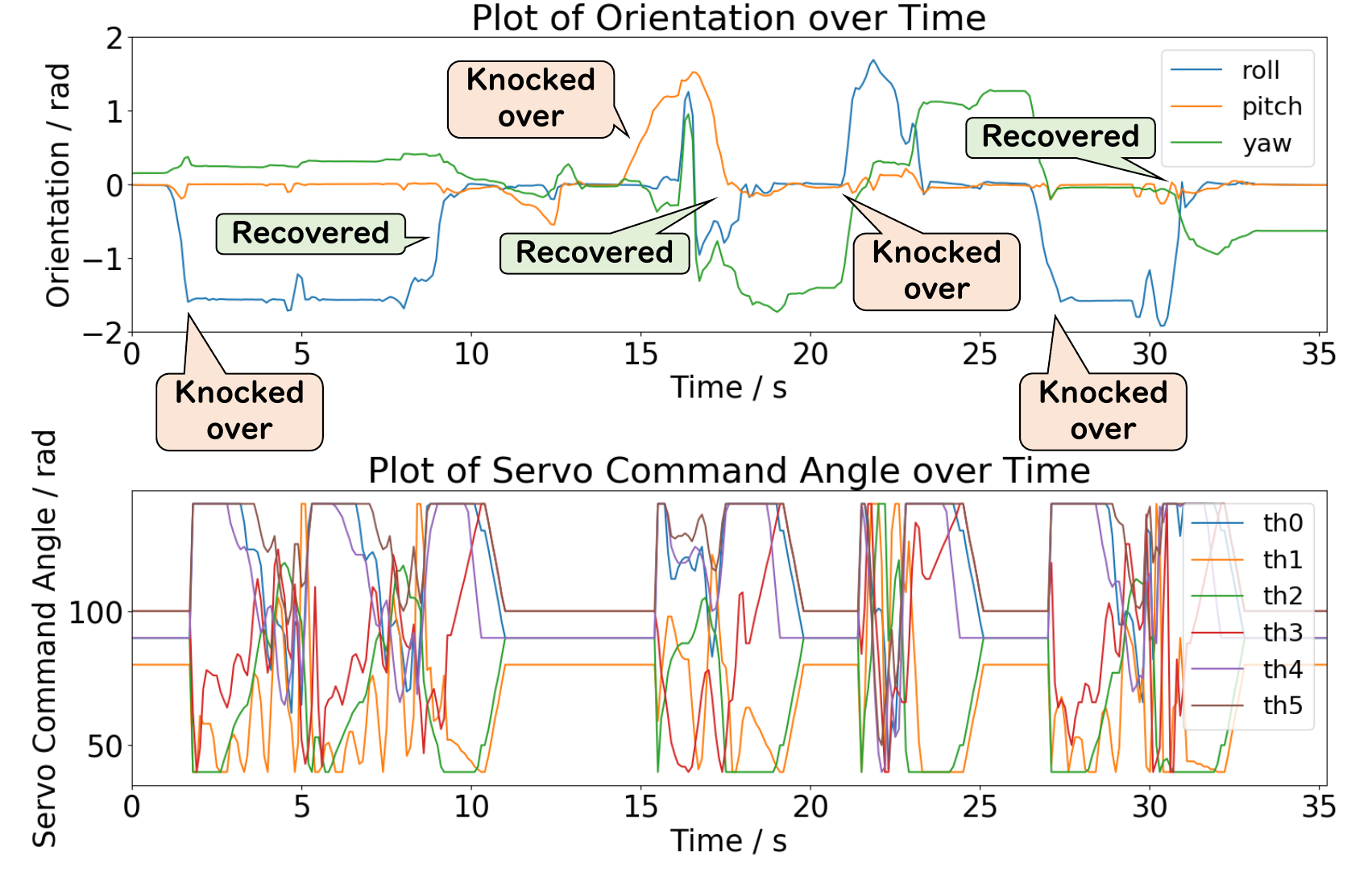}
    \caption{強化学習による様々な姿勢からの起き上がり.}
  	\label{figure:rl_stand_many_fig}
    \end{figure}

}%

\section{Discussion}
\switchlanguage%
{%

  The walking motion by connecting essential postures, involves the robot moving forward by putting out its 3 legs one by one in sequence.
  However, walking motion by reinforcement learning was the robot moving forward on three legs, vibrating the robot's entire body.
  The preference for vibrating motion in the walking motion by reinforcement learning can be attributed to the consideration of falling during walking. 
  In the walking style where the robot moves one leg at a time, it momentarily relies on two legs to support its body, then proceeds forward through a falling motion. 
  When raising one of the legs, or when the leg that has been lifted out re-enters the ground, the robot may tilt and fall, depending on the position of the legs.
  In contrast, the gait of reinforcement learning is to move forward with the three legs spread wide and stability maintained.
  Agents that have a falling gait are reset and not rewarded during learning, so they will have a non-falling gait after learning.
  Consequently, gait with reinforcement learning has a lower risk of falling than gait by connecting essential postures.
  Moreover, the pitch angle of the robot in walking by connecting essential postures remains approximately $\ang{0}$, 
  while it is approximately $\ang{35}$ in walking by reinforcement learning.
  This difference in pitch angles can be attributed to the asymmetry of the robot's body structure.
  In the case of the reinforcement learning approach, the robot moves forward while facing approximately $\ang{35}$,
  aligning the three legs symmetrically with its trajectory, effectively compensating for the body's asymmetry.

  Regarding the stand-up motion, both the connecting essential postures and reinforcement learning approaches utilize a reaction by tilting the seat 
  in the opposite direction of stand-up. 
  Additionally, in the reinforcement learning approach, a distinct behavior emerges during stand-up movements: 
  the robot attempts to stand-up by using the inertial force generated from swinging its legs.
  Furthermore, the robot achieves successful stand-up from various initial postures, such as the right side, back, or left side of the seat on the ground,
  guided by feedback from the IMU sensor within the inference model.
  It is thought that the robot has learned which leg should be kicked where in a certain posture to lead to the stand-up motion.

  In addition, for both movements, the reset conditions play a crucial role in the gait generation within the reinforcement learning process.
  Some movements resulted in the robot dragging its back and seat on the ground while moving forward during the walking motion.
  Other agents involved the seat facing upward but not fully stand-up.
  Such rewarding but unexpected movements could not be addressed by adjusting the reward's content or weight.
  The reset conditions effectively narrow the gait space for a robot with the incomplete and asymmetrical body structure, promoting the generation of the required gait.
  Conversely, relaxing the reset conditions could lead to increased chances of generating unexpected gaits. 
}%
{%
  歩行動作に関して，姿勢遷移による発見的な歩行動作は三本の脚を順番に１つずつ出していくのに対して，
  強化学習での歩行動作は三本脚でロボット全身を振動させながら前に進むものであった．
  強化学習においてそのような動きが発現したのには歩行動作中に転倒することに対する配慮であると考える．
  脚を１本ずつ前に出す歩き方は，あるタイミングにおいて２本脚で身体をささえる状況が存在し，そこから倒れるような動きを用いて前に進んでいく．
  ある１本の脚を上げる際，または出した脚が再び地面につく際において，脚の位置関係によってはロボットが傾いて倒れる可能性がある．
  一方で，強化学習の歩容は三本脚を大きく広げ，安定性を保ったまま全身する．
  学習時に転倒したものはリセットされ報酬が与えられないので，学習後は転倒しない歩容となる．
  一定速度で前進するほど前進性に優れているわけではない代わりに，転倒のリスクが姿勢遷移による発見的な歩行動作に比べて低い．
  また，歩行動作中のロボットのピッチ角が発見的な歩行動作ではおよそ$\ang{0}$であるのに対して，
  強化学習ではおよそ$\ang{35}$になっているのも違いである．
  これはロボット身体構造が非対称であることに起因していると考えられ，
  およそ$\ang{35}$を向きながら前に進んでいるということは，非対称な三本脚のロボットにとってみると進行軌跡に対して三本脚が対称に配置されることになり，
  身体構造の非対称性を克服した歩容であったと考えられる．
  
  起き上がり動作に関して，姿勢遷移による発見的な起き上がり動作が起き上がる方向と逆に座面を傾けた反動を用いるのと同様に，
  強化学習での起き上がり動作も一度，逆方向に座面を傾けるものが見られた．
  強化学習による起き上がり動作のみに見られた挙動として，脚を振り動かす慣性力によって起き上ろうとするものがあった．
  また，初期姿勢が座面の右側面を地面につけている状態だけでなく，背面や左側面を地面につけた状態から起き上がり可能であった．
  いずれも推論モデルに内在するIMUセンサによるフィードバックを活かした動作であり，あるロボットの姿勢においてどの脚でどこを蹴れば起き上がりの姿勢に繋がる
  のかが学習できていると考えれる．

  また，いずれの動作に対しても，強化学習においてはリセット条件が生成される歩容に重要であった．
  歩行動作では背中をや座面を地面に引き摺りながら前に進んでいく動きや，
  起き上がり動作では座面を上向きにするものの，そこから起き上がれない姿勢になるものがあったが，
  これらの報酬は高いが想定した動きではないものは，報酬の内容や重みを調節するだけでは解決されず，
  リセット条件をその動きがリセットされるように記述するのが効果的であった．
  複雑な身体構造を持つロボットにとって，その歩容の空間は広く，それをリセット条件によって狭めるのが
  要求する歩容を生成するのに有効であると考える．
  逆にリセット条件を緩めれば，思いもよらぬ歩容が生成される可能性が上がるだろう．
}%

\section{Conclusion} \label{sec:conclusion}
\switchlanguage%
{%
  In this study, we designed a chair-type tripedal low-rigidity robot with an asymmetric and imperfect body structure, and generated its gait using two methods: 
  connecting essential postures method and a reinforcement learning method.
  Both methods succeeded in generating walking and stand-up motions.
  In the walking motion, the gait generated by the reinforcement learning method was more stable than that by the connecting essential postures method.
  In the stand-up motion, both methods showed similar motions,
  but the gait with reinforcement learning potentially included sensor feedback and succeeded in stand-up motions from various postures.
  Even for robots with asymmetrical and imperfect body structures,
  it is possible to generate a gait by finding and connecting postures through trial and error or by reinforcement learning.
  While the connecting essential postures method can quickly produce a gait specific to the desired movement,
  the reinforcement learning method includes robust movements that take into account the robot's body structure,
  although there are difficulties in adjusting the conditions and parameters during learning.
}%
{%
  本研究では，非対称で不完全な身体構造をもつ椅子型三脚移動ロボットを設計し，姿勢接続による手法と強化学習による手法の2種類の手法を用いて
  その歩容を生成した．
  どちらの手法においても，歩行動作，起き上がり動作を実現することに成功した．
  歩行動作では強化学習による歩容において，姿勢接続手法よりもより安定した歩容が発現した．
  起き上がり動作では，どちらの手法も似た動きが見られたが，強化学習による歩容はセンサフィードバックを潜在的に内包しており，
  様々な姿勢からの起き上がり動作を成功させた．
  非対称で不完全な身体構造をもつロボットに対しても，試行錯誤によって姿勢を発見し，それを接続する歩用を生成することができ，
  強化学習によっても生成可能である．
  姿勢接続手法では望む動きに特化した歩容をすぐに出しやすいのに対して，
  強化学習手法では学習時の条件やパラメタ調節に難しさがあるもののロボットの身体構造を考慮するロバストな動きが含まれる．

}%

{
  \bibliographystyle{IEEEtran}
  \bibliography{bib}
}

\end{document}